\documentclass{article}[h=1.2in, l=2cm]
\usepackage[letterpaper,top=2cm,bottom=2cm,left=2cm,right=2cm,marginparwidth=1.75cm]{geometry}
\usepackage{amsmath}
\usepackage{amssymb}
\usepackage{natbib}
\usepackage{subcaption}
\usepackage{graphicx, wrapfig}
\usepackage{algorithm}
\usepackage{algorithmic}
\usepackage{colortbl}
\usepackage{xcolor}
\usepackage[dvipsnames]{xcolor}
\usepackage[x11names]{xcolor}
\usepackage{subcaption}
\usepackage[colorlinks=true, allcolors=blue]{hyperref}
\usepackage[table]{xcolor}
\usepackage{array}
\usepackage{booktabs}
\usepackage{multirow}
\usepackage{colortbl}
\usepackage{pgfplots}

\title{
Q-Net: Queue Length Estimation via Kalman-based Neural Networks}
\author{Ting Gao, Elvin Isufi, Winnie Daamen, Erik-Sander Smits, and Serge Hoogendoorn}
\date{}
\begin{document}

\maketitle
\begin{abstract}
Estimating queue lengths at signalized intersections is a long-standing challenge in traffic management. Partial observability of vehicle flows complicates this task despite the availability of two privacy-preserving data sources: (i) aggregated vehicle counts from loop detectors near stop lines, and (ii) aggregated floating car data (aFCD) that provide segment-wise average speed measurements. However, how to integrate these sources with differing spatial and temporal resolutions for queue length estimation is rather unclear. Addressing this question, we present Q-Net: a queue estimation framework built upon a state-space formulation. This design addresses key challenges in queue modeling, such as violations of traffic conservation assumptions. Q-Net follows the Kalman predict-update structure and maintains physical interpretability in both the state evolution and measurement models. Q-Net uses an AI-augmented Kalman filter to learn time-varying gain dynamics from data. The framework supports real-time implementation and improves spatial transferability by grouping aFCD measurements into fixed-size local groups, making the number of learnable parameters independent of section length. Evaluations on urban main roads in Rotterdam, the Netherlands, show that Q-Net outperforms baseline methods, tracks queue formation and dissipation accurately, and mitigates aFCD-induced delays. By combining data efficiency, interpretability, real-time applicability, and spatial transferability, Q-Net makes accurate queue length estimation possible without costly sensing infrastructure like cameras or radar. 
\end{abstract}

\section{Introduction}
Traffic queues raise operational, safety, and environmental concerns that require effective monitoring and control. Urban congestion reduces network throughput and limits agglomeration economies \citep{sweet2011does}. Long queues also increase accident likelihood \citep{pasquale2018optimal}, elevate noise levels \citep{chevallier2009improving}, contribute to driver stress and mental health issues \citep{nadrian2019sick}, and generate localized air pollution \citep{matzoros1992model}. Queue length estimation helps address these issues by enabling real-time adaptive signal control and, more broadly, supporting the development of queue-based control algorithms and network-wide optimization strategies \citep{li2021backpressure, hoogendoorn2016lessons}. However, accurate estimation remains difficult because of incomplete sensor coverage, limited data availability, unpredictable driver behavior, and complex network layouts.

Queue length estimation at signalized intersections relies on either Lagrangian or Eulerian input data. Lagrangian data are collected from mobile sensors moving with traffic flow, mainly GPS-equipped probe vehicles. Recent studies have used probe trajectories for queue estimation through methods ranging from shockwave theory to probabilistic likelihood-based approaches \citep{cheng2011cycle, car-07, zhao2019various, cheng2012exploratory, zhao2021maximum}. These methods often require prior knowledge of probe vehicle penetration rates and may raise privacy concerns because individual positions are shared with external monitoring systems \citep{rose2006mobile}. In contrast, Eulerian data capture traffic conditions over fixed spatial segments, coming from either stationary sensors or aggregated mobile sources. Computer vision systems can detect queue ends \citep{umair2021efficient, albiol2011video}, but their performance may degrade under poor lighting or bad weather. Radar systems can provide direct queue measurements, but are expensive to install and maintain \citep{bernas2018survey}.

This study focuses on two readily available Eulerian data sources: loop detector data and aggregated floating car data (aFCD). Loop detectors remain widely used because they provide accurate vehicle counts, occupancy, and speed measurements, and extensive detector infrastructure already exists in many cities \citep{muck2002using, car-01, geroliminis2011identification, car-05}. However, they provide point-based observations and cannot directly capture the queue length, especially under incomplete detector coverage. aFCD complements loop detectors by providing segment-level speed information. It is derived from anonymized mobile observations aggregated over road segments and time intervals, without requiring high penetration rates \citep{vanderloop2017validation, car-22, mobiliteitsscanFCD}. The key challenge is to integrate these heterogeneous data, with different spatial and temporal resolutions, into a unified and principled queue estimation framework.

\subsection{Literature review}
Queue length estimation at signalized intersections can be categorized into four approaches: input-output models \citep{webster1958traffic, sharma2007input}, shockwave theory \citep{lighthill1955kinematic, car-01}, speed-based methods \citep{car-23}, and filtering algorithms \citep{yin2018kalman}. 

{\em Input-output models} estimate queue length based on the conservation equation of cumulative traffic inflow and outflow at signalized sections. Early studies used vehicle counts from road detectors to characterize the queuing process \citep{webster1958traffic}, while later work incorporated real-time flow and occupancy measurements to adjust queue estimates \citep{vigos2008real}. 
While input-output models are straightforward, they do not capture the spatiotemporal dynamics of queues and are sensitive to measurement errors. Recent studies have improved them by integrating probe vehicle data, enabling real-time queue-tail estimation \citep{wang2021kalman, tang2023robust}; however, they require sufficient probe penetration.

{\em Shockwave theory} describes queue formation and dissipation, with shockwave speed indicating traffic state transitions. \citet{car-01} used loop detector data for cycle-wise queue estimation, later extended by \citet{car-05} to both undersaturated and oversaturated conditions. These methods identify occupancy breakpoints across signal phases, but such patterns can be obscured by noise and traffic heterogeneity. To address this, studies have fused probe vehicle data with detector measurements to estimate shockwave speeds and derive cycle-wise maximum and residual queues \citep{cai2014shock, wang2017shockwave}. \citet{car-14} proposed a hybrid method that applies input-output models when breakpoints are unreliable and shockwave analysis otherwise. Although shockwave-based models offer a theoretical basis for queue evolution, they assume uniform traffic conditions, which rarely hold in urban settings.

{\em Speed-based methods} estimate queue tails by analyzing speed patterns across road segments over time, typically using floating car data (FCD). Some methods detect queue tails from spatial speed drops in probe FCD \citep{car-23}. Because aFCD provides segment-level average speeds, \citet{car-22} generated synthetic probe trajectories using four-node quadratic interpolation and constructed iso-speed contours to estimate queue length. \citet{van2018estimating} combined aFCD with loop detector data to estimate vehicle accumulation. While such speed-based methods are intuitive and data-efficient, their accuracy depends heavily on data quality and congestion speed thresholds. More recently, \citet{car-08} avoided fixed speed thresholds by combining aFCD, loop detector, and traffic light event data within a decision-tree and RPROP-NN framework. However, such neural network-based methods require section-specific training and act as black boxes, limiting transferability and interpretability.

{\em Kalman filtering algorithms} use state-space models with a hidden state evolution equation and a measurement equation. The hidden state represents queue-related variables that evolve over time but are not directly observed, and different data sources can be integrated into either the evolution or measurement model. The recursive prediction-update structure of Kalman filtering enables real-time tracking of queue dynamics. Several studies have applied this framework to queue estimation. \citet{car-05} used an extended Kalman filter in which cycle-wise maximum and residual queues are modeled as states that remain static within each signal cycle, with a shockwave-based measurement model capturing phase breakpoints. \citet{car-07} used two independent extended Kalman filters to track queue formation and discharge waves, with measurements derived from shockwave location and speed. \citet{wang2021kalman} applied a standard Kalman filter in which the hidden state is the number of queued vehicles; its evolution model follows input-output theory, while its measurement model accounts for probe vehicle penetration. Similarly, \citet{car-09} used an input-output evolution model combined with a case-based reasoning measurement model. Despite their flexibility, traditional Kalman filters rely on linear or linearized system models and require full prior knowledge of system dynamics, including noise covariances. However, these assumptions are often unrealistic in urban traffic. For example, noise can be complex, unknown, and time-varying, which degrades model performance \citep{li2015kalman}. Queue dynamics are inherently nonlinear and heterogeneous \citep{ahmed2021examining, chauhan2021driving}, and become even more difficult to model under partial observability.

\subsection{Research gap and contribution}
Current queue length estimation approaches face three main limitations. First, many studies are developed and validated in well-controlled settings, overlooking partial observability of traffic flows. In urban networks, vehicles may enter or leave the study area through adjacent roads or unsignalized intersections without being monitored, due to sensor placement constraints or failures \citep{FHWA2006, shang2024hybrid}. Second, existing methods often depend strongly on specific data conditions. Loop-detector-based methods built on input-output or shockwave theory require accurate boundary counts or clear occupancy patterns, which are frequently unmet in practice. Meanwhile, aFCD, despite its low cost, wide coverage, and privacy advantages, remains underutilized for queue estimation. As a result, many methods generalize poorly across imperfect real-world sensor settings. Third, traditional and extended Kalman filters rely on linear or linearized models and predefined noise covariances, assumptions that are often unrealistic under dynamic intersection traffic, limiting adaptability and scalability.

To address these limitations, we propose Q-Net, a queue length estimation approach tailored to urban environments. Q-Net is built upon a state-space modeling approach integrated with an AI-augmented Kalman filtering framework. It handles partially observed traffic flows, uses widely available and privacy-preserving loop detector and aFCD inputs, and preserves interpretability through physically meaningful state evolution and measurement models. To overcome the limitations of traditional Kalman filters, Q-Net learns the Kalman gain dynamically using neural networks, allowing it to capture time-varying and nonlinear noise characteristics. This learning process builds on the KalmanNet principle \citep{revach2022kalmannet}, which has gained attention in signal processing through several recent variants \citep{buchnik2023latent, revach2023rtsnet, choi2023split}.

The main contributions of this paper are summarized as follows:
\begin{enumerate}
    \item \textbf{Interpretable state-space formulation under partial observability:} We develop a state-space model for queue length estimation using only loop detector and aFCD data. Unlike approaches that assume fully observed boundary flows, we explicitly address partial observability: our evolution model separates high-frequency queue dynamics from slow estimation drift caused by unmonitored flows. We also design a nonlinear measurement function that maps queue length to expected trajectory-based macroscopic speeds, enabling physics-informed fusion of data sources with different spatial and temporal resolutions.

    \item \textbf{Spatial transferability via AI-augmented filtering:} Q-Net balances modeling capacity and interpretability by using neural networks to compute the Kalman gain while maintaining physically meaningful states. To support spatial transferability, we introduce a local measurement grouping strategy considering that measurement covariances mainly depend on neighboring aFCD speeds. This makes the learnable parameter dimension independent of road section length, allowing Q-Net to be trained on one section and tested on another.

    \item \textbf{Real-world validation:} Q-Net is validated on main urban road sections in Rotterdam, the Netherlands. It outperforms several baseline models, transfers to adjacent road sections without fine-tuning, and maintains reliable performance under different traffic patterns and infrastructure configurations. Its online version further demonstrates potential for real-time adaptive traffic control. To our knowledge, this is the first application of AI-augmented Kalman filtering to queue length estimation.
\end{enumerate}

The remainder of this paper is organized as follows. Section~\ref{sec:problem_setup} defines the problem setup and challenges. Section~\ref{sec:KF_background} reviews Kalman filter fundamentals. Sections~\ref{sec:state-space model} and~\ref{sec:predict_update} present the Q-Net state-space formulation and filtering procedure, respectively. Section~\ref{sec:analysis_results} reports the Rotterdam case study. Finally, Section~\ref{sec:discussion} discusses limitations and future directions, and Section~\ref{sec:conclusion} concludes the paper.

\section{Problem setup and challenges}
\label{sec:problem_setup}
\begin{figure}[!t]
    \centering
    \includegraphics[width=0.8\linewidth]{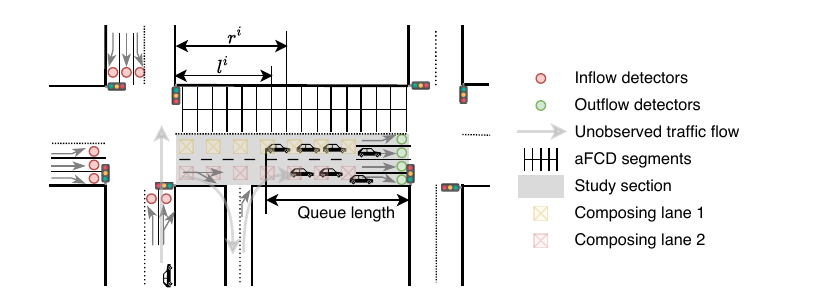}
     \caption{Layout of a signalized intersection and the section-wise queue length estimation setup. Segment $i$ extends from position $l^i$ to $r^i$, measured as distances from the stop line. Loop detectors provide inflow/outflow counts, while aFCD provides segment-level speed measurements. The figure also illustrates partial observability caused by multi-directional upstream lanes and unmonitored entering or leaving flows.}
    \label{fig:road_network}
\end{figure}

In this study, a \textit{lane} refers to a portion of the roadway used by a single line of vehicles. A \textit{section} is a group of lanes that share the same travel direction. As illustrated in Figure~\ref{fig:road_network}, a lane may accommodate multiple movement directions, and a section typically contains lanes that diverge into different movements at downstream intersections.

We estimate queue length at signalized intersections using indirect data sources, since direct queue measurement is often costly or impractical. The focus is section-wise queue length, defined as the maximum queuing distance across all lanes within a section. Consistent with \citet{ramezani2015queue}, a queue is not restricted to vehicles fully stopped at zero speed. Instead, we define it macroscopically as the spatial length of the section operating in a congested traffic state. In our formulation, the unqueued portion operates at free-flow speed ($v_f$), while the queued portion operates at a representative low speed ($v_j$). Since $v_j$ represents the average speed of the congested state, this definition includes both stopped vehicles and vehicles moving slowly in stop-and-go waves.

The problem involves three data sources, summarized in Table~\ref{tab:data}. Loop detectors installed near stop lines record vehicle passages, aggregated into 10-second vehicle counts. aFCD divides each road section into segments and provides average speed for each segment at 1-minute intervals. Radar sensors provide section-level queue length estimates at 10-second intervals. The radar data, hereafter referred to as the benchmark data, are used for model training and validation. Although treated as benchmark observations, they may still contain sensor-specific measurement noise.

Under this setup, queue length estimation faces three challenges. First, traffic flows are only partially observed. Figure~\ref{fig:road_network} illustrates two sources of unobserved vehicle movements: (i) vehicles from multi-directional upstream lanes that do not enter the downstream study section, and (ii) vehicles entering or leaving through unmonitored neighboring roads or facilities. Unmonitored entering flows are typically minor, whereas unmonitored exiting flows can be substantial, making direct aggregation of detector counts unreliable. Second, the data sources differ in spatial granularity and temporal resolution, as summarized in Table~\ref{tab:data}. Third, all data sources are subject to measurement noise: loop detectors may miscount due to calibration errors or occlusion, aFCD speeds may be affected by reporting delays and aggregation artifacts, and radar may mismeasure queues when sensor limits are exceeded.

These challenges motivate a state-space estimation framework that can combine indirect measurements with explicit uncertainty modeling. Before presenting Q-Net, the next section briefly reviews the Kalman filtering principles on which it builds.

\begin{table}[!t]
    \centering
    \caption{Overview of data sources and their characteristics}
    \begin{tabular}{c|c |c|c}
        \hline
        & \multicolumn{2}{c|}{Input Data} & \multicolumn{1}{c}{Benchmark Data} \\\hline
        Data Source & Loop Detector & aFCD & Radar \\
        Traffic Variable & Vehicle Count & Average Speed & Queue Length \\
        Spatial Coverage & Lane-level at stop lines (upstream and downstream) & Section segments & Section-level \\
        Temporal Resolution & 10 seconds & 60 seconds & 10 seconds \\
        Purpose & Model Input & Model Input & Evaluation \\
        \hline
    \end{tabular}
    \label{tab:data}
\end{table}

\begin{table}[!t]
    \centering
    \caption{Overview of notation, descriptions, and units.}
    \begin{tabular}{c|l|c}
        \hline
        \textbf{Notation} & \textbf{Description} &\textbf{Unit}\\
        \hline
        $t$ & Discrete time index & – \\
        $x_t$ & State variable at time $t$: section-level queue length & m \\
        $u_t$ & Control input at time $t$: estimated queue change  & m \\
        $\mathbf{y}_t$ & Measurement vector at time $t$ (aFCD speeds) & m/s \\
        $y_t^i$ & Speed measurement for $i$-th section segment at time $t$ & m/s \\
        $N$ & Number of aFCD segments within the section & – \\
        $w_t, Q_t$ & Process noise and its variance matrix & m, m² \\
        $\mathbf{v}_t, \mathbf{R}_t$ & Measurement noise and its covariance matrix& m/s, (m/s)² \\
        
        $\mathbf{h}$ & Measurement model function & – \\

        $L^q_t$ &Recovered queue length using only cumulative vehicle counts &m\\
        $\Sigma_t$ & Variance of state variable $x_t$& m² \\
        $\mathbf{S}_t$ & Covariance matrix of measurements $\mathbf{y}_t$& (m/s)² \\

        $\lambda_c$ & Unobserved traffic flow rate& vehicles/s \\
        $m$ & Number of lanes in the section & – \\
        $L$ & Length of the road section & m \\
        $A_t$ & Cumulative inflow (vehicle arrivals) count from upstream & vehicles \\
        $D_t$ & Cumulative outflow (vehicle departures) count from the section & vehicles \\
        $k_j, k_f$ & Jam density and free-flow density & vehicles/m \\
        $v_j, v_f$ & Jam speed and free-flow speed & m/s \\

        $Q_{\max}$ & Maximum reachable queue length for the section & m \\
        $\hat{x}_{t|t}$ & Posteriori estimate of $x_t$ given measurements up to time $t$ & m \\
        $\hat{y}_{t|t-1}$ & Predicted measurement at time $t$ based on state at time $t-1$ & m/s \\
        
        $l^i, r^i$ &Start and end positions of segment $i$ measured from the stop line & m \\
        $\Delta \tilde{x}_t$ & Forward evolution difference & m \\
        $\Delta \hat{x}_t$ & Forward update difference & m \\
        $\Delta \tilde{\mathbf{y}}_t$ & Measurement difference & m/s \\
        $\Delta {\mathbf{y}}_t$ & Innovation difference & m/s \\
        $\mathcal{K}_t(\theta)$&Kalman gain parameterized by $\theta$& – \\
        \hline
    \end{tabular}
    \label{tab:notation}
\end{table}
\section{Background in Kalman filtering}
\label{sec:KF_background}

State-space models describe how a system’s latent state and measurements (observations) evolve across time steps. Let $\mathbf{x}_t$ denote the latent state of the system at time $t$, $\mathbf{u}_t$ the control input, and $\mathbf{y}_t$ the observed measurement. The system is governed by a state evolution and a measurement (observation) model.

(1) State evolution describes how the state $\mathbf{x}_{t}$ evolves over time:
\begin{equation}
    \mathbf{x}_{t} = \mathbf{F}_t \mathbf{x}_{t-1} + \mathbf{G}_t \mathbf{u}_t + \mathbf{w}_t,\label{eq:back_model_1}
\end{equation}
where $\mathbf{F}_t$ is the state transition matrix, $\mathbf{G}_t$ is the control input matrix, and $\mathbf{w}_t \sim \mathcal{N}(\mathbf{0}, \mathbf{Q}_t)$ is zero-mean Gaussian process noise.

(2) Measurement model relates the observed measurement to the current state:
\begin{equation}
    \mathbf{y}_t = \mathbf{H}_t \mathbf{x}_t + \mathbf{v}_t,\label{eq:back_model_2}
\end{equation}
where $\mathbf{H}_t$ is the observation matrix and $\mathbf{v}_t \sim \mathcal{N}(\mathbf{0}, \mathbf{R}_t)$ is zero-mean Gaussian measurement noise. The noise $\mathbf{w}_t$ and $\mathbf{v}_t$ are mutually independent and also independent of both the states and the measurements. The initial state $\mathbf{x}_0$ is Gaussian with known mean $\hat{\mathbf{x}}_0$ and covariance matrix $\hat{\boldsymbol{\Sigma}}_0$.

The Kalman filter recursively estimates the \textit{a posteriori} distribution of the latent state $\mathbf{x}_t$ given all control inputs and measurements up to time $t$. Let $\boldsymbol{\Sigma}_t$ and $\mathbf{S}_t$ denote the state and measurement covariance matrices, respectively. At each time step, the filter alternates between prediction and update.

\paragraph{Prediction.} The filter first computes the \textit{a priori} state estimate $\hat{\mathbf{x}}_{t|t-1}$ and covariance $\boldsymbol{\Sigma}_{t|t-1}$ from the previous \textit{a posteriori} estimate $\hat{\mathbf{x}}_{t-1|t-1}$ and $\boldsymbol{\Sigma}_{t-1|t-1}$:

\begin{align}
    \hat{\mathbf{x}}_{t|t-1} &= \mathbf{F}_t \hat{\mathbf{x}}_{t-1|t-1} + \mathbf{G}_t \mathbf{u}_t, \label{eq:back_pred_1}\\
    \boldsymbol{\Sigma}_{t|t-1} &= \mathbf{F}_t \boldsymbol{\Sigma}_{t-1|t-1} \mathbf{F}_t^\top + \mathbf{Q}_t, \label{eq:back_pred_2}
\end{align}
where $(\cdot)^\top$ denotes the matrix transpose operation. The predicted measurement is:
\begin{align}
    \hat{\mathbf{y}}_{t|t-1} &= \mathbf{H}_t \hat{\mathbf{x}}_{t|t-1},\label{eq:back_pred_3} \\
    \mathbf{S}_{t|t-1} &= \mathbf{H}_t \boldsymbol{\Sigma}_{t|t-1} \mathbf{H}_t^\top + \mathbf{R}_t.\label{eq:back_pred_4}
\end{align}

\paragraph{Update.} Using the new measurement $\mathbf{y}_t$, the filter updates the state estimate:
\begin{align}
    \Delta\mathbf{y}_t&=\mathbf{y}_t-\hat{\mathbf{y}}_{t|t-1},\\
    \hat{\mathbf{x}}_{t|t} &=  \hat{\mathbf{x}}_{t|t-1}+\mathbf{\mathcal{K}}_t\Delta\mathbf{y}_t,\label{eq:back_update_1}\\
    \mathbf{\mathcal{K}}_t &=\mathbf{\Sigma}_{t|t-1}\mathbf{H}_t^\top\mathbf{S}_{t|t-1}^{-1},\\
    \mathbf{\Sigma}_{t|t} &= \mathbf{\Sigma}_{t|t-1}-\mathbf{\mathcal{K}}_t\mathbf{S}_{t|t-1}\mathbf{\mathcal{K}}_t^\top.\label{eq:back_update_2}
\end{align}

The innovation $\Delta\mathbf{y}_t$ is the prediction error between the observed measurement $\mathbf{y}_t$ and its one-step-ahead prediction $\hat{\mathbf{y}}_{t|t-1}$. The Kalman gain $\mathbf{\mathcal{K}}_t$ weights this innovation and, under a linear state-space model with fixed Gaussian noise, minimizes the mean squared error of the state estimate. This recursive prediction-update structure is the basis for online state estimation. For nonlinear systems, the Extended Kalman Filter applies the same principle by locally linearizing the state and measurement models using Jacobian matrices \citep{maybeck1982stochastic}.

\section{Q-Net state-space formulation}
\label{sec:state-space model}

Let $x_t \in \mathbb{R}$ be the section-level queue length at time step $t$, $u_t \in \mathbb{R}$ the queue change as the control input, and $\mathbf{y}_t = [y_t^1, y_t^2, \dots, y_t^N]^\top \in \mathbb{R}^N$ the segment-wise aFCD speed measurements over $N$ section segments. Q-Net adopts the following state-space model:
\begin{align}
x_{t} &= x_{t-1} + u_{t} + w_{t}, \label{eq:evolution}\\
\mathbf{y}_t &= \mathbf{h}(x_t) + \mathbf{v}_t. \label{eq:measurement}
\end{align}

The evolution equation \eqref{eq:evolution} models the current queue length as the previous queue length plus the queue change derived from \textit{loop detector counts} and process noise. The measurement equation \eqref{eq:measurement} relates the queue length to \textit{aFCD speeds} through a nonlinear mapping $\mathbf{h}(\cdot)$. Here, $w_t \sim \mathcal{N}(0, Q_t)$ denotes Gaussian process noise, and $\mathbf{v}_t \sim \mathcal{N}(\mathbf{0}, \mathbf{R}_t)$ denotes Gaussian measurement noise. The derivation of $u_t$ is given in Subsection~\ref{sec:control_input}, and the measurement function $\mathbf{h}(\cdot)$ is detailed in Subsection~\ref{sec:measurement_model}.

We do not treat noise covariances as fixed or known; instead, $Q_t$ and $\mathbf{R}_t$ are modeled as time-varying quantities, whose effects on the Kalman gain are learned from data. The state variance and measurement covariance are denoted respectively by $\Sigma_t$ and $\mathbf{S}_t$. Table~\ref{tab:notation} summarizes the notation used throughout the paper.

\subsection{Control queue change input}
\label{sec:control_input}
This subsection derives the control input $u_t$ in Equation~\eqref{eq:evolution}. The approach first reconstructs a queue-length proxy from \textit{cumulative vehicle counts} and then uses its temporal differences as queue changes.

Let $A_t$ and $D_t$ denote cumulative vehicle inflow and outflow up to time $t$, and let $L^q_t$ denote the reconstructed queue length. The control input is obtained as:
\begin{align}
    L^q_t &= \mathcal{C}(A_t-D_t),\\
    \bar{L}^q_t &=\mathcal{F}(L^q_t),\\
    u_t &= \bar{L}^q_t-\bar{L}^q_{t-1},
\end{align} 
where $\mathcal{C}$ reconstructs $L^q_t$ from vehicle count data while accounting for unobserved flows, $\mathcal{F}$ applies a Fourier band-pass filter to produce the smoothed queue length $\bar{L}^q_t$, and $u_t$ is computed as the temporal difference of $\bar{L}^q_t$. Applying correction and filtering before differencing yields a noise-robust queue-change input for Equation~\eqref{eq:evolution}.

\noindent\textbf{Step 1: Queue length recovery $\mathcal{C}(A_t-D_t)$.} Let $\lambda_c$ be the unobserved net traffic flow rate, defined as the net number of vehicles leaving the section per unit time without passing through the designated outflow detector, partially offset by unobserved minor inflow. We model a section of length $L$ with $m$ lanes and consider approximately balanced queues across lanes. The queued portion operates at jam density $k_j$ and speed $v_j$, while the remaining portion operates at free-flow density $k_f$ and speed $v_f$. Applying macroscopic vehicle conservation gives:
\begin{equation}
    m L^q_t k_j + m(L-L^q_t) k_f = A_t - D_t - \lambda_c t.
    \label{eq:conservation}
\end{equation}
The left-hand side represents the total amount of vehicles in the queued and free-flow portions of the section, while the right-hand side represents net accumulation adjusted for unobserved flows. Solving for $L^q_t$ yields:
\begin{equation}
L^q_t = \frac{A_t - D_t - mL k_f - \lambda_c t}{m(k_j - k_f)}.
\label{eq:x_experession}
\end{equation} 

\noindent\textbf{Step 1.1: Estimate $\lambda_c$.} Since $A_t-D_t$ is obtained from loop detector counts, the remaining unknown governing the evolution of $L^q_t$ is $\lambda_c$. We estimate it using boundary conditions at the beginning $t_0$ (e.g., 6:00 AM) and end $t_T$ (e.g., 8:00 PM) of each daily analysis period. This relies on the physical assumption that queues are negligible ($L^q_t \approx 0$) at these pre-peak ($t_0$) and post-peak ($t_T$) times. Applying these conditions yields:
$\lambda_c \approx \frac{(A_{t_T}-D_{t_T})-(A_{t_0}-D_{t_0})}{t_T-t_0}$.

\begin{figure}[!t]
    \centering
    \begin{subfigure}{0.48\linewidth}
        \centering
        \includegraphics[width=\linewidth]{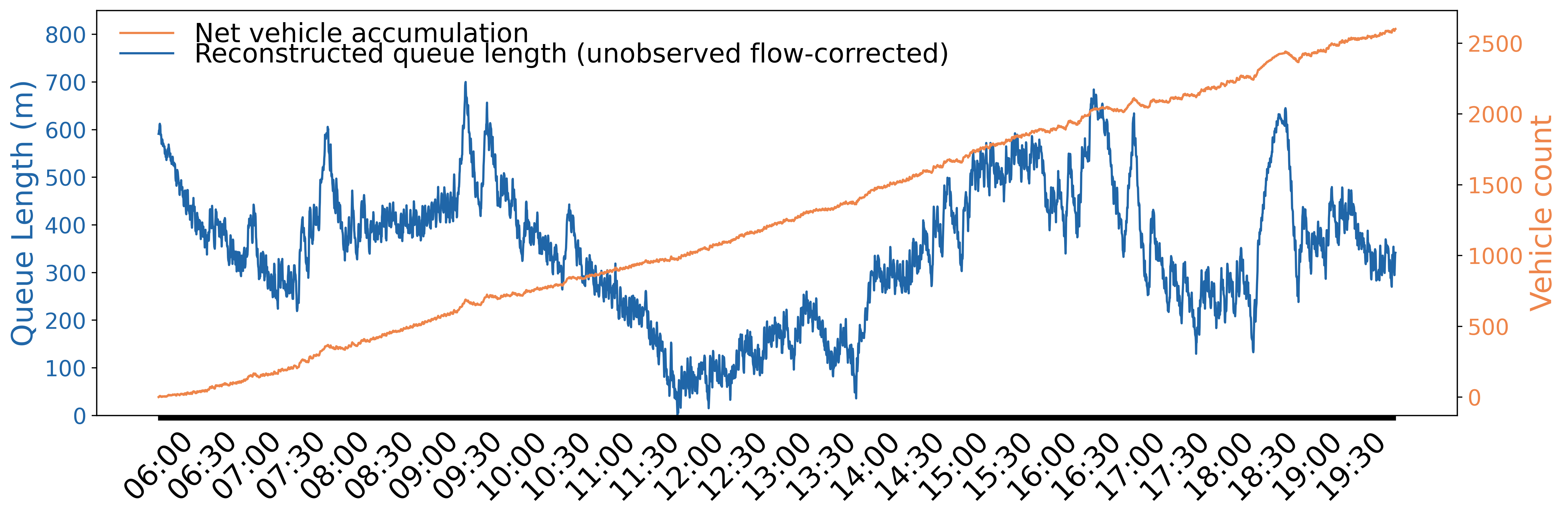}
        \caption{Reconstructed queue length $L^q_t$ is obtained by removing the unobserved traffic flow and applying an affine transformation to align with the observed queue range.}
        \label{fig:control_input1}
    \end{subfigure}
    \hfill
    \begin{subfigure}{0.48\textwidth}
        \centering
        \includegraphics[width=\textwidth]{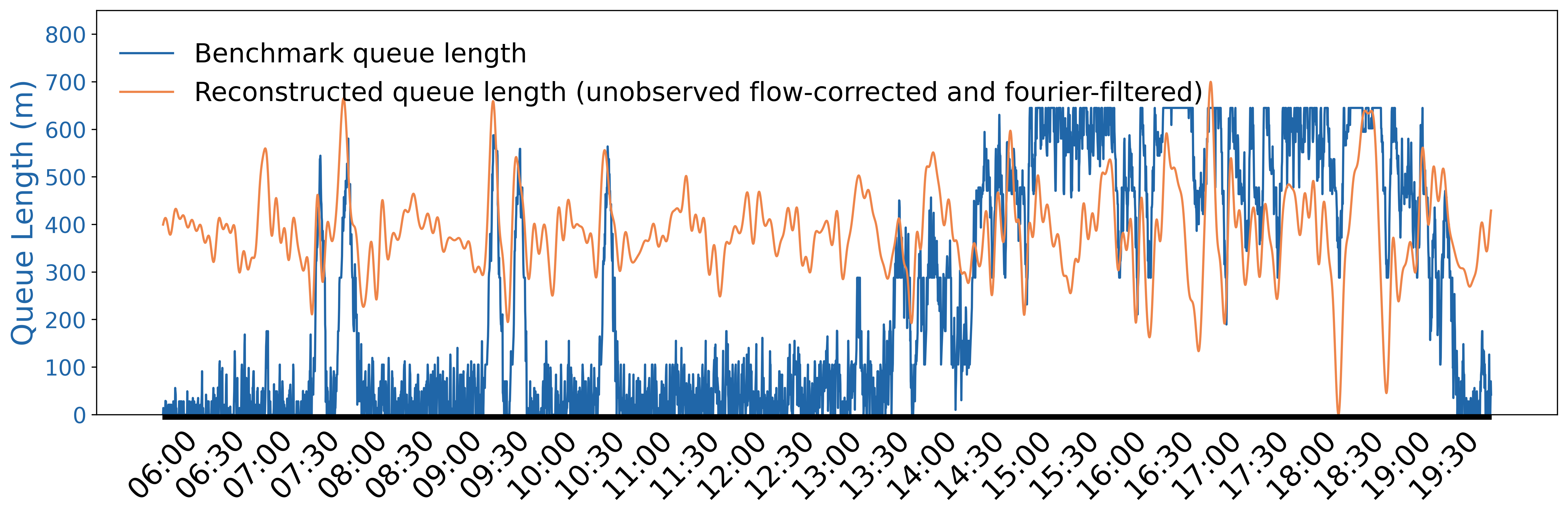}
        \caption{Fourier filter suppresses low-frequency drift and high-frequency noise, producing a smoothed queue profile $\bar{L}^q_t$ suitable for control input $u_t$.}
        \label{fig:control_input2}
    \end{subfigure}
    \caption{Queue estimation from net vehicle accumulation under unobserved traffic flow conditions for section \texttt{N1-IN} in Rotterdam on 2023-11-10.}
    \label{fig:control_input}
\end{figure}

\noindent\textbf{Step 1.2: Estimate $L^q_t$.} Equation~\eqref{eq:x_experession} shows that $L^q_t$ is an affine transformation of $(A_t-D_t-\lambda_c t)$. Rather than estimating $k_j$ and $k_f$, which can be time-varying and site-specific, we directly shift and scale this signal to the empirically observed queue range $[0, Q_{\max}]$, where $Q_{\max}$ is the section-specific maximum reachable queue length. Figure~\ref{fig:control_input1} shows the resulting reconstructed queue length $L^q_t$.

\medskip\noindent\textbf{Step 2: Fourier band-pass filtration $\mathcal{F}(L^q_t)$.} As shown in Figure \ref{fig:control_input1}, the reconstructed queue length $L^q_t$ may contain both low-frequency drift and high-frequency noise. The drift can arise from time-varying unobserved flow rate $\lambda_c$, such as changes in turning ratios, while high-frequency noise may originate from detector errors. We apply a Fourier-based band-pass filter $\mathcal{F}(\cdot)$ to suppress both components and retain medium-frequency variations that better represent queue buildup and dissipation. Figure~\ref{fig:control_input2} shows the filtered profile $\bar{L}^q_t=\mathcal{F}(L^q_t)$, from which $u_t$ is computed. Because $\bar{L}_t^q$ is used only to derive the control input, \textit{its absolute magnitude is less important than its temporal variation.} Thus, even if the reconstructed queue levels have similar average values in the morning and afternoon, this has limited impact as long as the relative queue changes are preserved.

\medskip\noindent\textbf{Remark on oversaturation and spillback.} The queue evolution model relies on measured boundary flows rather than expected phase capacities. If downstream congestion limits green-phase discharge, the downstream detector records the reduced outflow; if spillback blocks upstream entry, the upstream detector records zero inflow. When the section is fully congested and lacks a free-flow portion, the aFCD measurement model reports aggregated speeds near the congested speed $v_j$. Since Q-Net models the section as an aggregated storage unit, partial blocking across movements, such as a left-turn queue blocking adjacent through traffic, is not modeled explicitly at the lane level but is reflected in the aggregate measured boundary flows. The adaptive process noise captures the increased uncertainty under oversaturated conditions, while aFCD measurements help correct possible estimation drift.

\subsection{Measurement function}
\label{sec:measurement_model}
The measurement function maps the section-level queue length $x_t$ to the expected aFCD speed on each segment. We use the two-state traffic representation introduced in Subsection~\ref{sec:control_input}: the queued portion operates at congested speed $v_j$, while the unqueued portion operates at free-flow speed $v_f$. Let segment $i$ extend from position $l^i$ to $r^i$, measured as distances from the stop line with $r^i>l^i$. At time $t$, the expected speed for segment $i$ is:
\begin{equation}
    [h(x_t)]_i=
    \left\{
    \begin{array}{l@{\quad}l@{\quad}l}
    v_f & x_t\leq l^i & \text{(segment is entirely unqueued)},\\
    \frac{r^i-l^i}{\frac{x_t-l^i}{v_j}+\frac{r^i-x_t}{v_f}} & l^i<x_t\leq r^i & \text{(segment is partially queued)},\\
    v_j & r^i< x_t & \text{(segment is entirely queued)}.
    \end{array}
    \right.
\label{eq:h_expression}
\end{equation}
For partially queued segments, where the queue boundary lies within the segment, the expected speed is computed as segment length divided by total travel time, with travel time accumulated over the queued and unqueued portions. The noisy measurement is given by $y_t^i = [h(x_t)]_i + v_t^i$, consistent with Equation~\eqref{eq:measurement}. 

The characteristic speeds $v_j$ and $v_f$ are estimated as the two peak modes of the historical aFCD speed distribution. This empirical calibration aligns the measurement function with section-specific aFCD observations, while short-term cycle-to-cycle fluctuations are treated as stochastic measurement noise.

The speed $[h(x_t)]_i$ is defined from a macroscopic, travel-time-based perspective, consistent with aFCD aggregation. aFCD providers aggregate observations from passing probe vehicles over time intervals, so reported segment speeds represent aggregated traversal speeds rather than instantaneous arithmetic averages of all vehicles present on the segment \citep{ndw_fcd_2026}. Accordingly, Equation~\eqref{eq:h_expression} computes expected speed from the travel time needed to traverse the free-flow and queued portions. Because $v_j$ is typically much lower than $v_f$, vehicles spend more time in the congested portion, so the resulting speed is naturally weighted toward $v_j$ as the queue grows.

\begin{figure}[!t]
    \centering
    \includegraphics[width=0.8\linewidth]{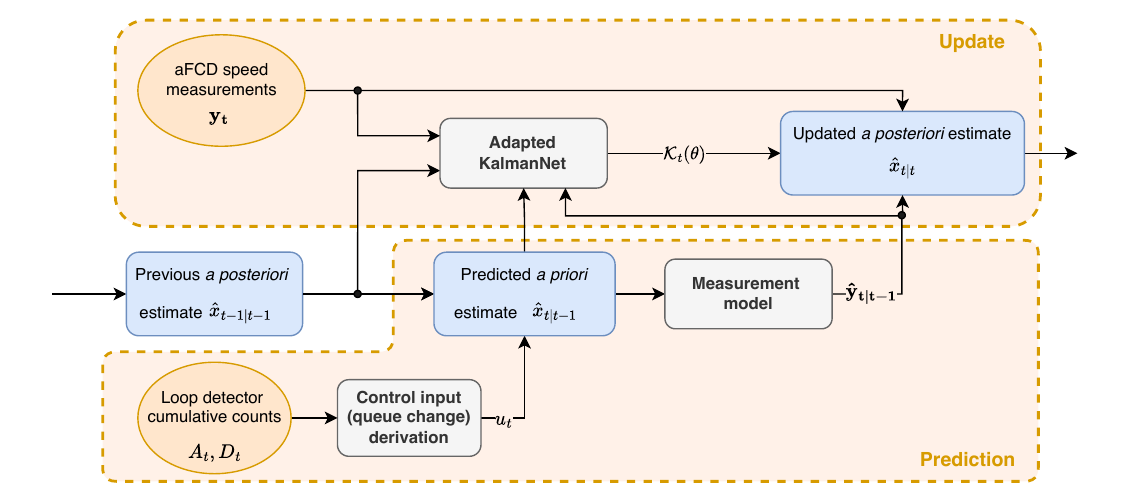}
    \caption{
    Overview of the Q-Net filtering framework at time $t$. The predict step uses cumulative vehicle counts $A_t,D_t$ to derive the control input (queue change) $u_t$ and predict the queue length $\hat{x}_{t|t-1}$ and expected aFCD speeds $\hat{\mathbf{y}}_{t|t-1}$. The update step compares $\hat{\mathbf{y}}_{t|t-1}$ with observed aFCD speeds $\mathbf{y}_t$ and uses the adapted KalmanNet to compute the Kalman gain $\mathcal{K}_t$ and update the queue estimate $\hat{x}_{t|t}$.}
    \label{fig:method_overview}
\end{figure}

\section{Queue-length estimation framework}
\label{sec:predict_update} 
This section presents the Q-Net filtering procedure. As shown in Figure~\ref{fig:method_overview} and Algorithm~\ref{alg:qnet}, Q-Net follows a Kalman-style predict-update loop.
\begin{algorithm}[!t]
\caption{Queue-length Estimation Algorithm}
\label{alg:qnet}
\textbf{Input:} Initial state $\hat{x}_{0|0}$, control inputs $\{u_t\}$, measurements $\{\mathbf{y}_t\}$, maximum queue length $Q_{\max}$, parameters $\theta$ \\
\textbf{Output:} Queue length estimates $\{\hat{x}_{t|t}\}$

\textbf{For} $t = 1, 2, \ldots, T$:

\hspace*{1em}\textbf{Predict:} 
    
\hspace*{3em}$\hat{x}_{t|t-1} = \mathbb{P}_{[0,Q_{\max}]} (\hat{x}_{t-1|t-1}+u_t)$ $\leftarrow$ \textit{Equation}~\eqref{eq:predict_function}
        
\hspace*{3em}$\hat{\mathbf{y}}_{t|t-1} = \mathbf{h}(\hat{x}_{t|t-1})$ $\leftarrow$ \textit{Equation}~\eqref{eq:h_expression},\eqref{eq:predict_y}

\hspace*{1em}\textbf{Update:}
    
\hspace*{3em}$\text{context}_t=(\Delta \tilde{x}_t, \Delta \hat{x}_t, \Delta \tilde{\mathbf{y}}_t, \Delta {\mathbf{y}}_t)$ $\leftarrow$ \textit{Equation}~\eqref{eq:context1},\eqref{eq:context2},\eqref{eq:context3},\eqref{eq:context4}
        
\hspace*{3em}$\mathcal{K}_t = \text{NeuralNetwork}(\theta, \text{context}_t)$
        
\hspace*{3em}$\hat{x}_{t|t} = \mathbb{P}_{[0,Q_{\max}]} (\hat{x}_{t|t-1} + \mathcal{K}_t (\mathbf{y}_t - \hat{\mathbf{y}}_{t|t-1}))$ $\leftarrow$ \textit{Equation}~\eqref{eq:kalmannet_update}
\end{algorithm}

\medskip\noindent\textbf{Predict.} The predict step estimates the \textit{a priori} queue length $\hat{x}_{t|t-1}$ from the previous \textit{a posteriori} estimate $\hat{x}_{t-1|t-1}$ and the control input $u_t$. It then predicts the expected aFCD speeds through the measurement function $\mathbf{h}(\cdot)$:
\begin{align}
    \hat{x}_{t|t-1} &=\mathbb{P}_{[0,Q_{\max}]}  (\hat{x}_{t-1|t-1}+u_{t}), \label{eq:predict_function}\\
    \hat{\mathbf{y}}_{t|t-1} &= \mathbf{h}(\hat{x}_{t|t-1}), \label{eq:predict_y}
\end{align}
where $\mathbb{P}_{[0,Q_{\max}]}(\cdot)$ projects the state estimate onto the physically feasible range $[0,Q_{\max}]$, preventing negative queues or queues exceeding the section capacity. This projection introduces a nonlinear constraint, rendering it sub-optimal under classical Kalman filter theory. Nevertheless, Q-Net learns the Kalman gain directly from data rather than relying on linear-Gaussian assumptions. The neural network is expected to adapt to the nonlinearity introduced by the projection during training.

\medskip\noindent\textbf{Update.} The update step corrects the predicted state using the innovation $\mathbf{y}_t-\hat{\mathbf{y}}_{t|t-1}$ weighted by a learned Kalman gain $\mathcal{K}_t(\theta)$. Q-Net learns $\mathcal{K}_t(\theta)$ from data following the KalmanNet principle. This is important because our state-space model is nonlinear (Section~\ref{sec:state-space model}) and the process and measurement uncertainties vary over time, as illustrated by the different noise patterns in the morning and afternoon in Figure~\ref{fig:control_input}. The projection in Equation~\eqref{eq:kalmannet_update} again enforces physically meaningful estimates and is consistent with constrained filtering practice \citep{simon2010kalman, prakash2014recursive}:
\begin{equation}
    \hat{x}_{t|t} = \mathbb{P}_{[0,Q_{\max}]}(\hat{x}_{t|t-1}+\mathcal{K}_t(\theta) (\mathbf{y}_t-\mathbf{\hat{y}}_{t|t-1})).
    \label{eq:kalmannet_update}
\end{equation}

Loop detector counts are available every 10 seconds, while aFCD speeds arrive at a coarser 60s resolution. Since aFCD represents an aggregated one-minute speed rather than an instantaneous snapshot, we use a sample-and-hold strategy: when a new aFCD measurement arrives, it replaces the previous value; otherwise, the most recent value is retained. Although the held measurement remains constant within the one-minute interval, the innovation remains time-varying because $\hat{\mathbf{y}}_{t|t-1}$ changes every 10 seconds as $\hat{x}_{t|t-1}$ in Equation~\eqref{eq:predict_y} is driven by loop detector inputs. Because aFCD speeds are temporally aggregated whereas $\mathbf{h}(x_t)$ maps an instantaneous queue state to an expected speed, the resulting temporal mismatch is treated as part of the measurement uncertainty. \textit{Since aFCD is used as a lower-frequency correction signal, while high-resolution loop detector counts drive state propagation in the prediction step, residual aFCD timing errors may still introduce some delay but are less dominant in determining short-term queue dynamics in the offline experiments.} The separate issue of causal aFCD availability, which can further increase the effective temporal mismatch, is examined through the online variants in Section~\ref{sec:real_time}.

\subsection{Learned Kalman gain}
\label{sec:kalmannet}
Equation~\eqref{eq:kalmannet_update} is the key departure from traditional Kalman filtering: Q-Net learns the Kalman gain from data instead of computing it analytically. Following \citet{revach2022kalmannet}, four temporal features are used to represent the statistical relationship between the state-space model and the gain:
\begin{align}
    \text{The }&\textit{forward evolution difference: }\quad\Delta \tilde{x}_t = \hat{x}_{t|t}-\hat{x}_{t-1|t-1}; \label{eq:context1} \\
    \text{The }&\textit{forward update difference: }\quad\Delta \hat{x}_t = \hat{x}_{t|t}-\hat{x}_{t|t-1}; \label{eq:context2} \\
    \text{The }&\textit{measurement difference: }\quad\Delta \tilde{\mathbf{y}}_t = \mathbf{y}_t-\mathbf{y}_{t-1}; \label{eq:context3}\\
    \text{The }&\textit{innovation difference: }\quad\Delta {\mathbf{y}}_t = \mathbf{y}_t-\mathbf{\hat{y}}_{t|t-1}. \label{eq:context4}
\end{align}
The \textit{forward evolution difference} $\Delta \tilde{x}_t $ captures the change between consecutive \textit{a posteriori} state estimates, while the \textit{forward update difference} $\Delta \hat{x}_t$ captures the correction from the \textit{a priori} to the \textit{a posteriori} estimate. Since $\hat{x}_{t|t}$ is unavailable before the Kalman gain is computed, the previous values $\Delta \tilde{x}_{t-1}$ and $\Delta \hat{x}_{t-1}$ are used during inference. In this setup, the pair ($\Delta \tilde{x}_{t-1}$, \textit{measurement difference} $\Delta \tilde{\mathbf{y}}_t$) captures state and measurement evolution, while $\Delta \hat{x}_{t-1}$ and the \textit{innovation difference} $\Delta {\mathbf{y}}_t$ reflect the current correction and innovation uncertainty.

These inputs are fed into the KalmanNet architecture, which uses Gated Recurrent Units (GRUs) to learn time-varying uncertainty representations. The GRUs control how past information impacts the current prediction, mirroring the recursive nature of the traditional Kalman gain calculation (Equations \eqref{eq:back_pred_1}-\eqref{eq:back_update_2}). The resulting parametric Kalman gain, detailed in Appendix~\ref{sec:appendix_Kalmannet}, is computed as:
\begin{equation}
    \mathcal{K}_t =\text{KalmanNet}(\Delta \tilde{x}_t, \Delta \hat{x}_t, \Delta\tilde{\mathbf{y}}_t, \Delta {\mathbf{y}}_t).
\end{equation}

\paragraph{Training pipeline.} Q-Net is trained end-to-end to learn the Kalman gain. Given a sequence of $T$ time steps, the loss is the Root Mean Square Error (RMSE) between the estimated queue length $\hat{x}_{t|t}$ and the radar benchmark $x_t$:

\begin{equation}
\mathcal{L}=\sqrt{\frac{1}{T}\sum_{t=0}^{T-1}(x_t-\hat{x}_{t|t})^2}
\end{equation}
During training, the projections in Equations~\eqref{eq:predict_function} and \eqref{eq:kalmannet_update} are disabled to ensure uninterrupted gradient flow. They are applied during validation and testing to enforce physical queue length bounds.

\subsection{Adaptation for spatial transferability}
\label{sec:kalmannet_adapted}
\begin{figure}[!t]
    \centering
    \includegraphics[width=0.8\linewidth]{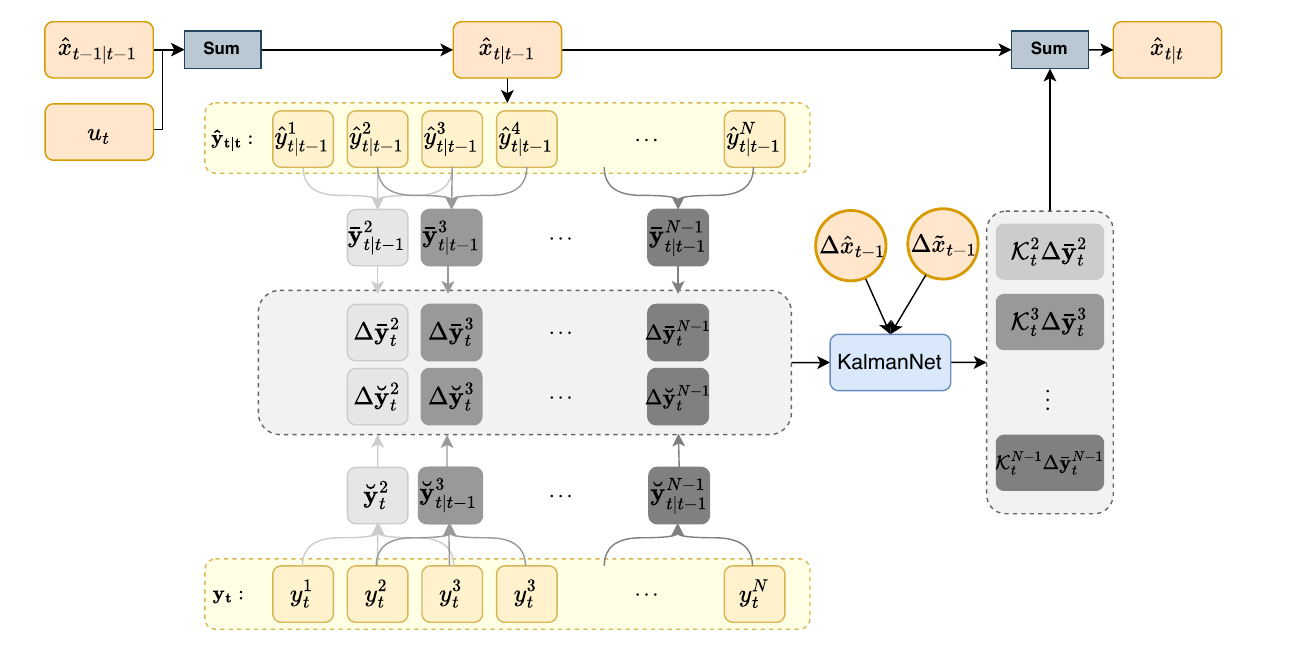}
    \caption{Local measurement grouping strategy for transferable Kalman gain estimation. aFCD measurements are grouped into fixed-size vectors consisting of the previous, current, and next segments. Each group is processed by the same KalmanNet function to obtain a local gain, and the final state update is obtained by summing all local updates.}
    \label{fig:grouped_prediction}
\end{figure}
    
A segment is the spatial unit over which aFCD aggregates speed measurements. Since benchmark queue data are difficult and costly to obtain, a model trained on one instrumented section should ideally be transferable to other sections. However, standard KalmanNet requires a fixed-dimensional measurement vector determined by the road section’s aFCD segmentation, which varies across sections. This mismatch motivates our spatial transferability adaptation. To address it, Q-Net uses fixed-size local measurement groups, enabling the same trained model to be applied to sections with different lengths or segmentations.

The grouping is based on the assumption that measurement covariances depend only on the segment itself and its immediate neighbors (previous and next segments), which is justified by the short 10-second filtering interval (benchmark data frequency). As shown in Figure \ref{fig:grouped_prediction}, for each intermediate segment $1<i<N$, we form:
\begin{itemize}
    \item The \textit{grouped measurement difference} $\Delta \breve{\mathbf{y}}^i_t = [{y}_t^{i-1}-y_{t-1}^{i-1}, {y}_t^{i}-y_{t-1}^{i}, {y}_t^{i+1}-y_{t-1}^{i+1}]^{\top}$;
    \item The \textit{grouped innovation difference} $\Delta {\mathbf{\bar{y}}}^i_t = [{y}_t^{i-1}-{\hat{y}}_{t|t-1}^{i-1}, {y}_t^{i}-{\hat{y}}_{t|t-1}^{i}, {y}_t^{i+1}-{\hat{y}}_{t|t-1}^{i+1}]^{\top}$.
\end{itemize}
Boundary segments ($i=1$ and $i=N$) are excluded because queue dynamics are mainly reflected by intermediate segments. The first segment near the stop line can be erroneous due to localized dissipation patterns, while the last segment adds limited information since intermediate segments repeatedly confirm the presence of a long queue.

Each grouped feature vector is fed to \textit{the same KalmanNet} function. Instead of learning one global gain vector in Equation~\eqref{eq:back_update_1}, Q-Net learns multiple, local gain vectors $\mathcal{K}^i_t$ for each group $i$:

\begin{align}
    \mathcal{K}_t^i &=\text{KalmanNet}(\Delta \tilde{x}_t, \Delta \hat{x}_t, \Delta \breve{\mathbf{y}}^i_t, \Delta {\mathbf{\bar{y}}}^i_t), &\text{where }\mathcal{K}^i_t\in\mathbb{R}^{1\times 3},\\
    \hat{x}_{t|t} &= \hat{x}_{t|t-1} + \sum_{i=2}^{N-1} \mathcal{K}^i_t \Delta\bar{\mathbf{y}}^i_t, &\text{where } \Delta {\mathbf{\bar{y}}}^i_t \in\mathbb{R}^3. \label{eq:group_KF}
\end{align}
Since a single measurement innovation $\Delta y_t^j$ appears in three groups ($j-1$, $j$, and $j+1$), its total effective gain is implicitly the sum of components from three adjacent group gains ($\mathcal{K}^{j-1}_t$, $\mathcal{K}^{j}_t$, and $\mathcal{K}^{j+1}_t$). While this is not a globally optimal update in the classical sense, this design localizes the gain calculation. This localization makes the architecture independent of the total section segment number ($N$) and thereby achieves spatial transferability.

\section{Numerical results}

\label{sec:analysis_results}

In this section, we evaluate Q-Net in terms of temporal transferability (Section~\ref{sec:performance}), spatial transferability (Section~\ref{sec:transfer}), online capability (Section~\ref{sec:real_time}), and component contributions (Section~\ref{sec:abaltion_study}). Training is supervised and relies on benchmark queue data. Sections~\ref{sec:performance} and~\ref{sec:transfer} report offline ex-post evaluations using full-day data. In these evaluations, each aFCD timestamp is shifted backward by one minute to align the reported speed with the preceding aggregation interval. Section~\ref{sec:real_time} assesses online performance.

\begin{figure}[!t]
    \centering
    \includegraphics[width=\linewidth]{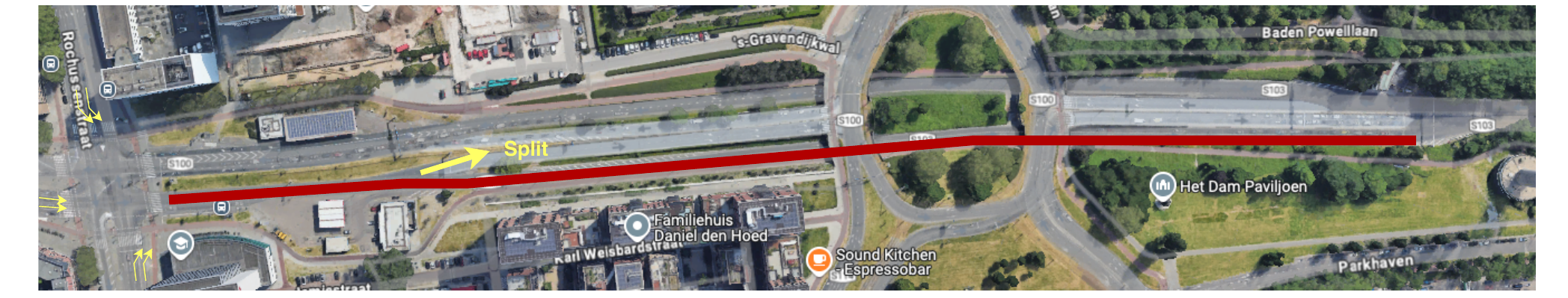}
    \caption{Satellite view of \texttt{N1-IN}. The study section is highlighted in dark red. Downstream, part of the traffic flow exits toward a grade-separated interchange connected to an open bridge, leading to substantial unobserved outflow.}
    
    \label{fig:N1-IN}
\end{figure}

\medskip\noindent\textbf{Case study.} We consider a real-world case study in Rotterdam, the Netherlands, a dense urban road network with recurrent congestion. The dataset covers 16 days from 6:00 AM to 8:00 PM, including 12 weekdays and 4 weekend days. We exclude sections with insufficient data availability due to contractual restrictions or substantial missing values, as the evaluation focuses on queue estimation rather than data imputation. Among the remaining candidates, we select \texttt{N1-IN} as the main study section (Figure~\ref{fig:N1-IN}). This major urban section connects directly to the Maastunnel, an important entry point to Rotterdam and a critical focus of traffic management. It is challenging because considerable unobserved traffic exits toward a ring interchange. Missing aFCD values are imputed using the most recent available value within the same segment.

\medskip\noindent\textbf{Baselines.} 
We compare Q-Net with relevant baseline methods. Many existing approaches are not directly comparable because they rely on assumptions incompatible with our setting. For example, loop-detector-based methods estimate cycle-wise maximum queues \citep{car-01, car-05, car-06, car-14} or assume fully observed and conserved traffic flow \citep{car-09, car-10, car-17, car-20}. These assumptions do not hold in our study section, which has limited loop coverage and substantial unobserved traffic flow. In addition, cycle-wise methods typically estimate per-lane or per-signal-group maximum queues, whereas our benchmark data provide section-level queue lengths every 10 seconds.

Given these constraints, we select six baselines. The first two use only aFCD: \textbf{OSD} \citep{car-22}, which detects queue ends from observed speed drops using fixed upstream and downstream thresholds of 16 km/h; and \textbf{ISC} \citep{car-23}, which constructs synthetic trajectories from aFCD and estimates queues using iso-speed contours with thresholds from 14 to 22 km/h. The third, \textbf{Tree-RPROP} \citep{car-08}, is a multisource baseline combining aFCD, loop detector, and traffic light event data through a decision tree and a resilient propagation neural network. We adopt the same setup as the original paper. Finally, we include three supervised data-driven baselines using the same input features as Q-Net: Long Short-Term Memory networks (\textbf{LSTM}) \citep{hochreiter1997long}, Gaussian Processes (\textbf{GP}) \citep{rasmussen2006gaussian}, and eXtreme Gradient Boosting (\textbf{XGBoost}) \citep{chen2016xgboost}. Their hyperparameters are selected by grid search.

\begin{figure}[!t]
    \centering
    \includegraphics[width=0.8\linewidth]{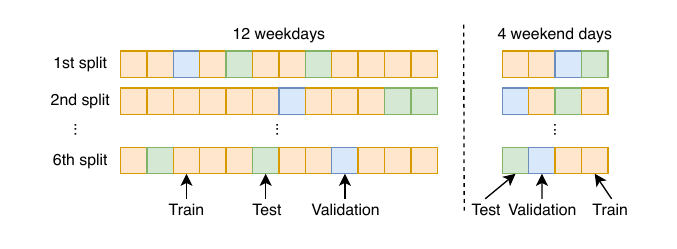}
    \caption{Six-fold data split: for each split, weekday and weekend samples are randomly assigned to the training, validation, and test sets.}
    \label{fig:data_split}
\end{figure}

\medskip\noindent\textbf{Q-Net implementation and hyperparameters.} 
The Fourier band-pass filter is tuned on the validation set. The lower and upper cutoff frequencies are set to 1/360 and 1/30 cycles per step. With a 10-second system time step, this preserves signal periods between 300s (30 steps) and 3600s (360 steps). The KalmanNet implementation follows \citet{revach2022kalmannet}\footnote{\url{https://github.com/KalmanNet/KalmanNet_TSP}}. Q-Net is trained with the Adam optimizer using a learning rate of 0.001. It contains 952 trainable parameters, and each training epoch takes approximately 40 seconds on an Apple M2 Pro chip.

\medskip\noindent\textbf{Training strategy.} At 10-second resolution, each daily series contains 5,040 time steps. To reduce accumulated prediction errors during the early stage of training, we divide each sequence into 10-minute windows. The state is initialized to zero for the first window of each day. For subsequent windows, the initial state is set to the \textit{a posteriori} estimate from the previous window, mimicking practical deployment where the true state is unobserved.

\medskip\noindent\textbf{Evaluation protocol.} Due to the limited data, we use six independent train-validation-test splits for robust evaluation. For each split, the model is trained and tested independently: 11 days are randomly selected for training, 2 for validation, and 3 for testing, while preserving a mix of weekdays and weekends (Figure~\ref{fig:data_split}). This ensures that every day appears in the test set at least once. The day-wise splits are reported in Appendix~\ref{appendix:data_split}.

Models with random initialization are trained with five random seeds. Performance is evaluated using Root Mean Square Error (RMSE), Mean Absolute Error (MAE), and Mean Absolute Percentage Error (MAPE). Metrics are computed by concatenating predictions and benchmark queue lengths across all test splits. To avoid inflated relative errors near zero, MAPE is computed only when the benchmark queue length exceeds 10 meters, covering 71.73\% of the test set. Morning and afternoon peaks are defined as 07:00--09:00 and 16:00--18:00, following Dutch Ministry of Infrastructure guidelines \citep{mobiliteitsscanFCD}.

\subsection{Performance}
\label{sec:performance}

This subsection evaluates temporal transferability, where Q-Net is trained and tested on the same section using different time periods. This setting applies when benchmark sensors, such as radar, can be temporarily installed to collect training data and then removed to reduce long-term sensing costs.

\begin{figure}[!t]
    \centering
    \includegraphics[width=\linewidth]{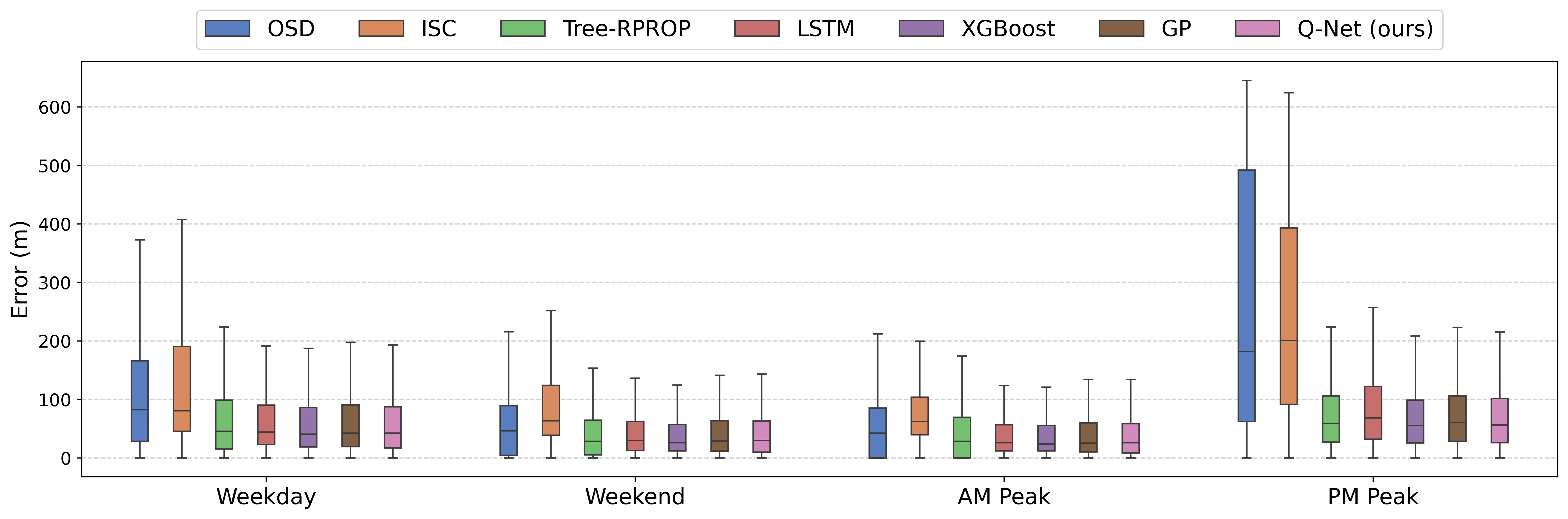}
    \caption{Box plots of absolute estimation errors for baseline models and Q-Net for section \texttt{N1-IN}.}
    \label{fig:error_box}
\end{figure}

\medskip\noindent\textbf{Quantitative analysis.} Table~\ref{tab:N1-IN} reports model performance, and Figure~\ref{fig:error_box} shows absolute error distributions. Q-Net achieves the best all-day RMSE and MAE and remains competitive across peak periods. OSD and ISC perform worst because they rely directly on aFCD and are therefore sensitive to speed measurement errors. Q-Net also improves over Tree-RPROP, despite Tree-RPROP using additional data sources. Although standard data-driven regression models, especially XGBoost, achieve comparable time-averaged metrics, these aggregate values do not fully reveal transient estimation errors. The qualitative analysis below further examines these differences.

\begin{table}[!t]
    \centering
    \caption{Performance comparison of the proposed and baseline models when trained and tested on \texttt{N1-IN}.}
    \begin{tabular}{c|ccc|ccc|ccc}
        \hline
         &\multicolumn{3}{c|}{\textbf{All day}}&\multicolumn{3}{c|}{\textbf{Morning peak}}&\multicolumn{3}{c}{\textbf{Afternoon peak}}\\
        &\textbf{RMSE} & \textbf{MAE} & \textbf{MAPE} &\textbf{RMSE} & \textbf{MAE} & \textbf{MAPE} &\textbf{RMSE} & \textbf{MAE} & \textbf{MAPE} \\\hline
OSD&200.17&121.56&95.48&122.94&71.24&118.53&349.54&269.32&58.02\\
ISC&191.18&131.22&94.03&165.84&108.07&135.33&297.51&241.49&50.3\\
Tree-RPROP&114.25&67.96&65.11&111.17&58.5&91.74&101.37&75.49&19.43\\
GP&92.56&61.26&51.79&84.88&50.11&67.99&93.25&73.32&18.5\\
LSTM &89.32&61.04&49.35&80.85&49.11&66.32&103.32&82.54&18.16\\
XGBoost &87.81&58.17&48.52&82.17&48.56&63.58&85.82&67.71&16.96\\
Q-Net (ours)&86.77&58.14&56.22&78.62&47.62&74.64&88.36&69.25&17.41\\\hline
    \end{tabular}
    \label{tab:N1-IN}
\end{table}

\begin{figure}[p]
    \centering
    \begin{subfigure}[b]{\textwidth}
        \centering
        \includegraphics[width=\textwidth]{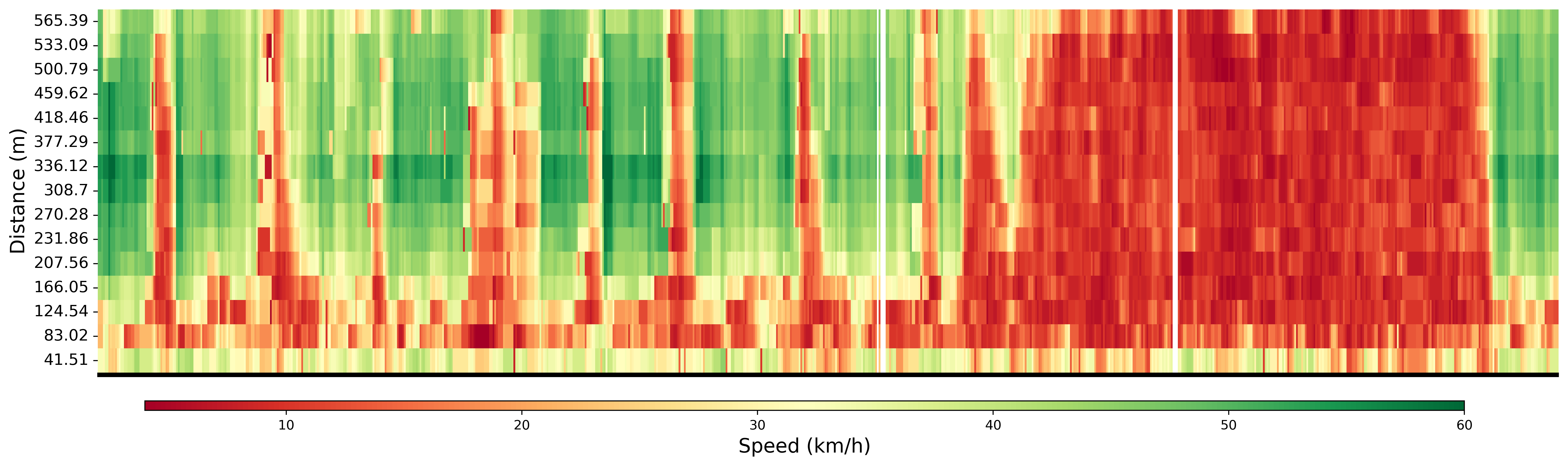}
        \caption{The aFCD.}
        \label{fig:fcd}
    \end{subfigure}
    \begin{subfigure}[b]{\textwidth}
        \centering
        \includegraphics[width=\textwidth]{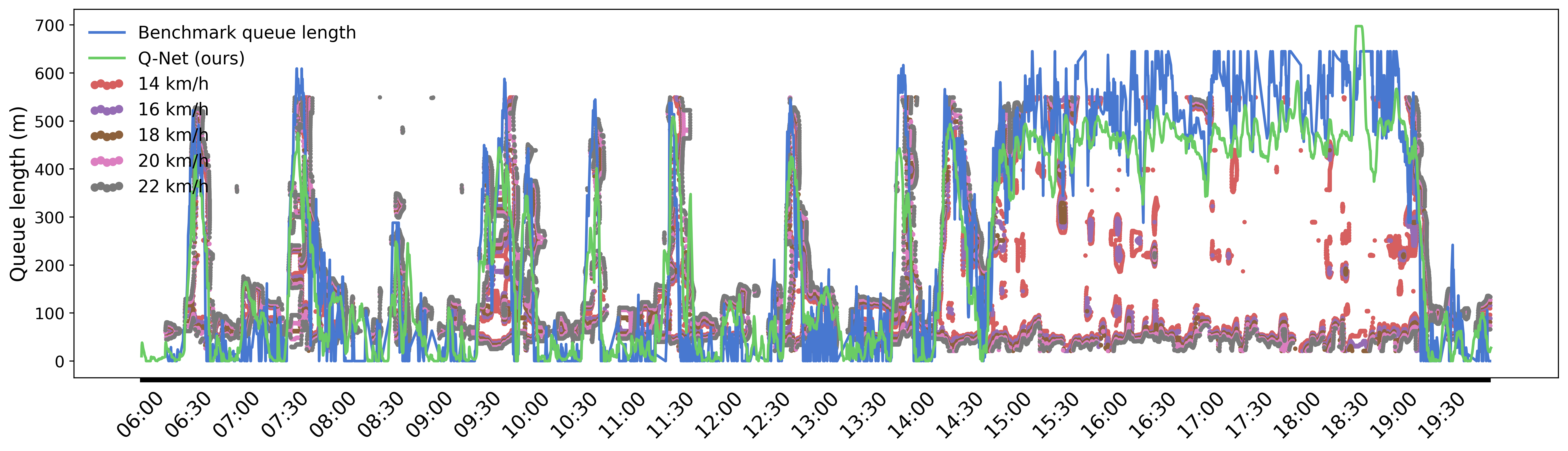}
        \caption{Q-Net vs ISC.}
    \end{subfigure}
    \begin{subfigure}[b]{\textwidth}
        \centering
        \includegraphics[width=\textwidth]{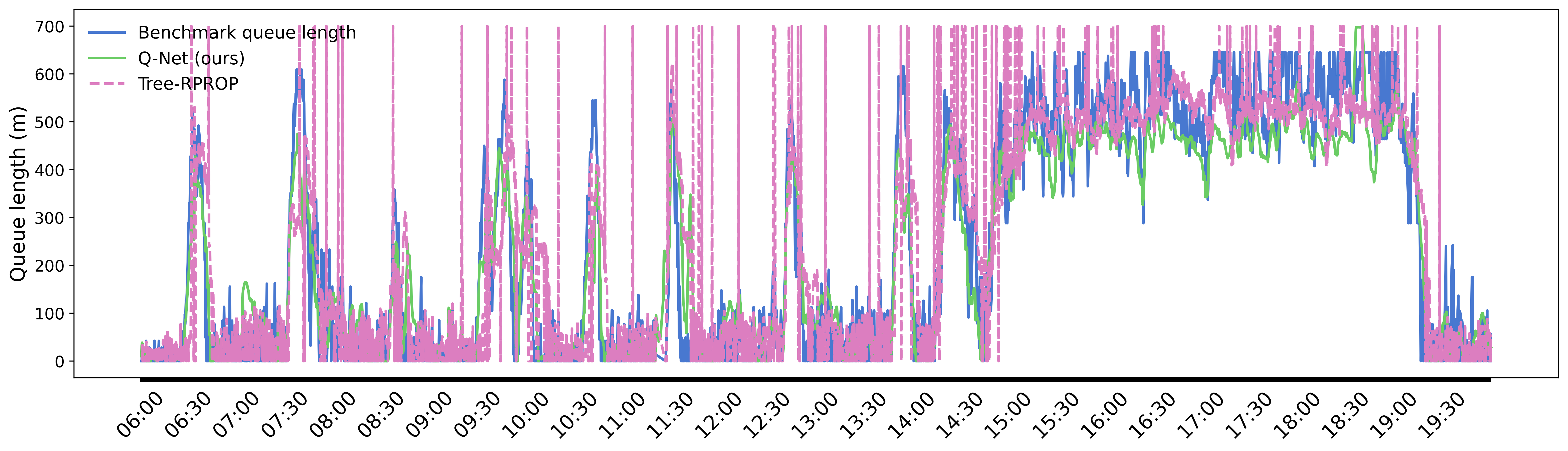}
        \caption{Q-Net vs Tree-RPROP.}
    \end{subfigure}
    \begin{subfigure}[b]{\textwidth}
        \centering
        \begin{minipage}[b]{0.76\textwidth}
            \centering
            \includegraphics[width=\textwidth]{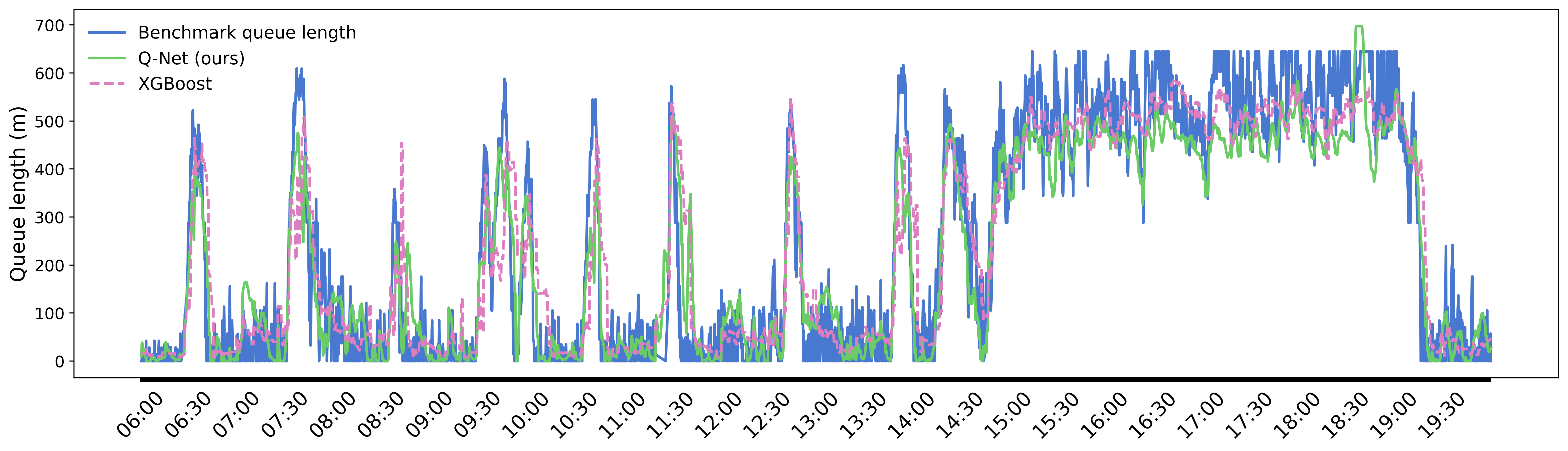}
        \end{minipage}
        \hfill
        \begin{minipage}[b]{0.23\textwidth}
            \centering
            \includegraphics[width=\textwidth]{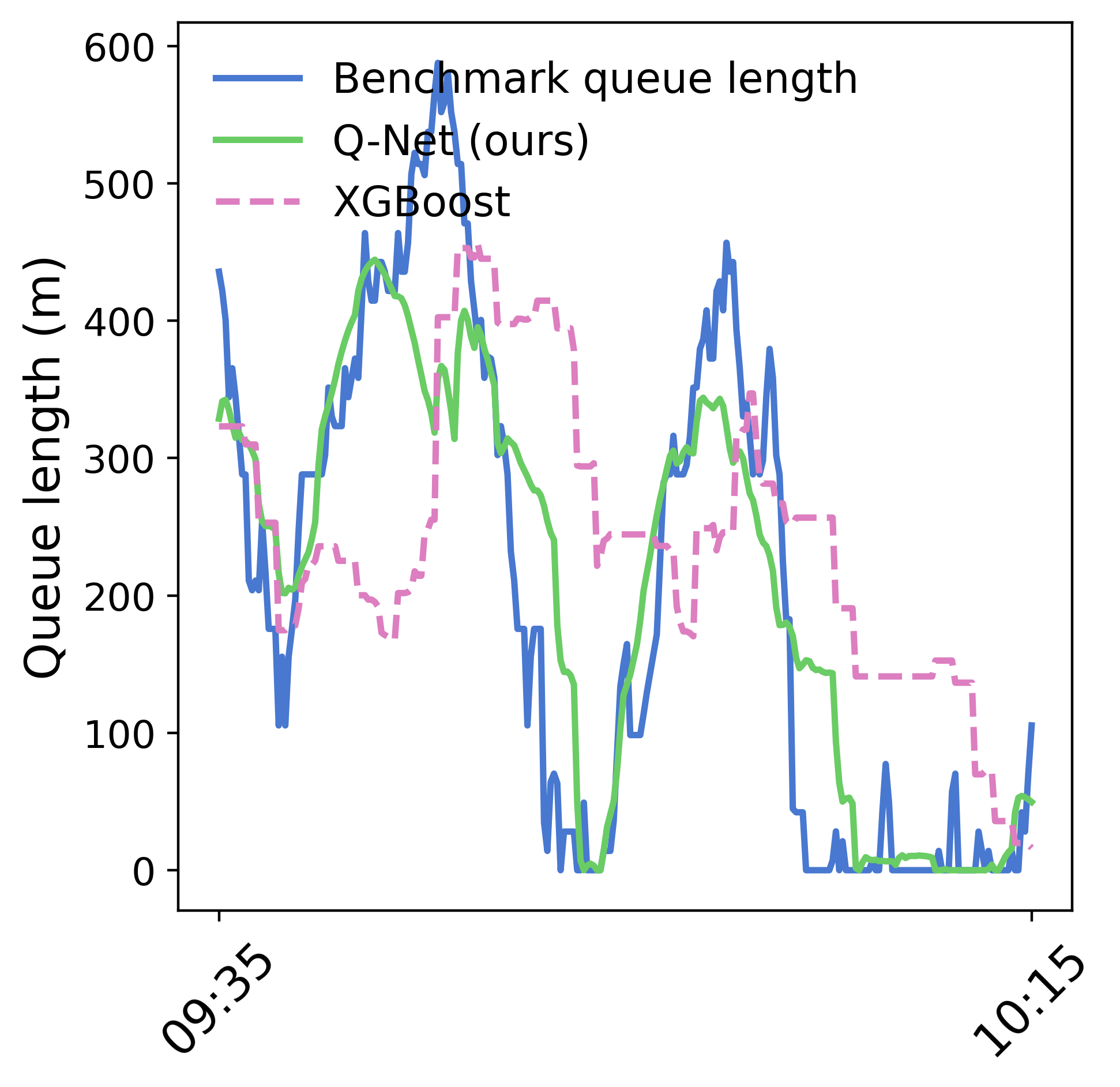}
        \end{minipage}
        \caption{Q-Net vs XGBoost. Left: full-day estimation. Right: zoomed-in comparison during rapid queue variation.}
        \label{fig:XGBoost_vs_QNet}
    \end{subfigure}

    \caption{The aFCD visualization and queue length estimation comparison among selective baselines (ISC, Tree-RPROP, and XGBoost), and our proposed method for section \texttt{N1-IN} on 2023-11-14. Results are generated using the model trained with random seed 42.}
    \label{fig:visual_comparison}
\end{figure}

A notable observation from Table~\ref{tab:N1-IN} is the higher morning-peak MAPE across all methods. The benchmark data indicate that queues do not consistently extend across the full morning peak window. Since MAPE is relative to the benchmark value, shorter queues produce larger percentage errors. Figure~\ref{fig:error_box} also shows smaller absolute errors during weekends and morning peaks, when queues are generally less pronounced and therefore easier to estimate.

\medskip\noindent\textbf{Qualitative analysis.} Figure~\ref{fig:visual_comparison} shows a visual case study for 2023-11-14. For clarity, we plot the best-performing model from each baseline category: ISC, Tree-RPROP, and XGBoost. ISC suffers from discontinuities and may generate multiple queue estimates at a single time step. Tree-RPROP produces many unrealistic maximum queues because it relies on a fixed loop detector occupancy threshold under noisy conditions. Although XGBoost achieves competitive time-averaged metrics, it exhibits clear estimation lag during rapid queue formation and dissipation, as shown in the zoomed comparison in Figure~\ref{fig:XGBoost_vs_QNet}. Without an explicit queue-evolution structure, purely data-driven models tend to respond reactively to observed input patterns. In contrast, Q-Net uses loop-detector-derived queue changes to drive the prediction step, enabling it to track rapid buildup and dissipation more promptly, while aFCD measurements correct slower estimation drift in the update step. This prediction-correction structure reduces temporal lag while maintaining smooth estimates.

The baseline errors in Figure~\ref{fig:visual_comparison} are associated with aFCD noise and temporal lag. Figure~\ref{fig:fcd} shows several aFCD artifacts: (a) erroneous high speeds within the queue during congestion, such as 15:00-19:00, which can place the queue end incorrectly in the middle of the section; (b) high speeds near the stop line, likely due to sparse probe coverage or aggregation artifacts, leading to near-zero queue estimates despite congestion; and (c) isolated low-speed patches that cause inconsistent queue-end detection. In addition, despite temporal alignment, aFCD can still exhibit delay due to probe sparsity, segment aggregation, and vendor-side smoothing \citep{ndw_fcd_2026}. Q-Net is less affected by these errors because the learned Kalman gain adapts to changing model uncertainty, while loop-detector-derived queue changes provide a more responsive prediction of queue dynamics. During 13:30-14:00, for example, aFCD shows delayed queue formation, but Q-Net still captures the queue onset because loop detector cumulative counts respond more promptly to queue changes.

\subsection{Transferability}
\label{sec:transfer}
This subsection evaluates spatial transferability, where Q-Net is trained on one section and tested on others without fine-tuning. This setting is relevant when radar benchmark data are available for one instrumented section but not for every target section. The grouping strategy in Subsection~\ref{sec:kalmannet_adapted} enables this transferability.

\begin{figure}[!t]
    \centering
    \includegraphics[width=\linewidth]{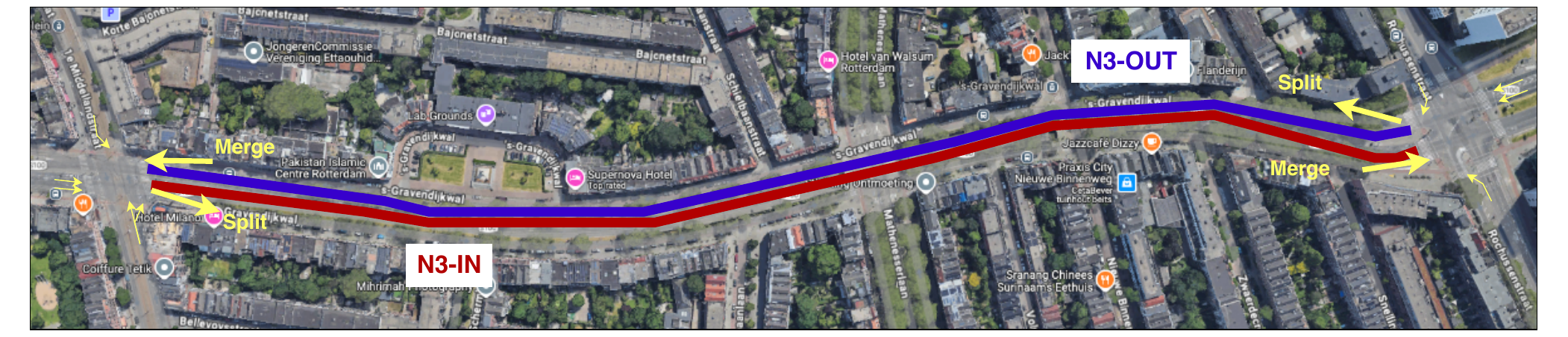}
    \caption{Satellite view of \texttt{N3-IN} and \texttt{N3-OUT}. In both cases, the downstream area divides into an underground study section and an above-ground branch, which merge at the downstream intersection.}
    \label{fig:AFM_N3_INUIT}
\end{figure}

We introduce two additional Rotterdam sections adjacent to \texttt{N1-IN}: \texttt{N3-IN} and \texttt{N3-OUT} (Figure~\ref{fig:AFM_N3_INUIT}). Both are underground sections with an above-ground branch that diverges after the upstream intersection and merges back downstream. In \texttt{N3-IN}, combined through/right-turn lanes at the upstream intersection create additional unobserved flow in the studied direction. We train Q-Net separately on \texttt{N1-IN}, \texttt{N3-IN}, and \texttt{N3-OUT}, and evaluate each trained model on all three sections.

\begin{figure}[!t]
    \centering
    \includegraphics[width=\linewidth]{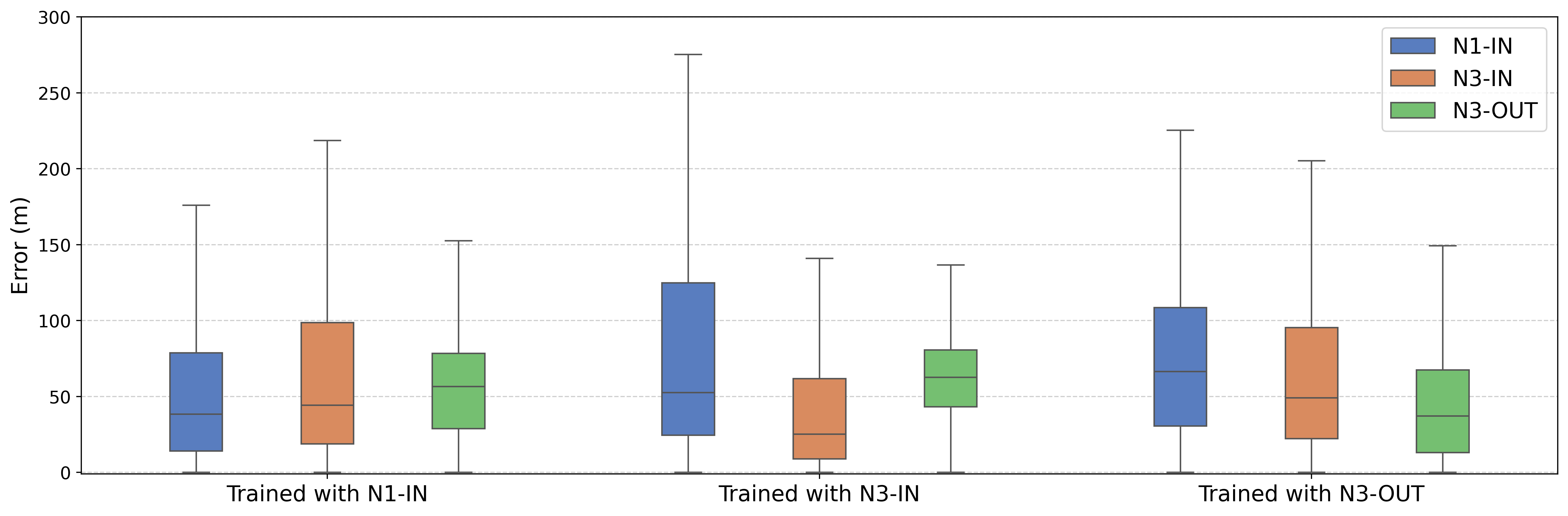}
    \caption{Box plots of absolute estimation errors across all test datasets using different training sections.}
    \label{fig:error_box_transferability}
\end{figure}

Figure~\ref{fig:error_box_transferability} shows the performance for all training-testing combinations. As expected, each section performs best when the model is trained on itself, indicating that Q-Net captures location-specific traffic dynamics. Interestingly, the model trained on \texttt{N1-IN} transfers better to \texttt{N3-IN} and \texttt{N3-OUT} than vice versa. Although \texttt{N1-IN} has a simpler infrastructure layout, its unobserved upstream split exposes the model to uncertainty patterns that transfer effectively to the two N3 sections. By contrast, \texttt{N3-IN} and \texttt{N3-OUT} contain additional complexity: above-ground branches diverge and rejoin downstream, introducing unobserved traffic at both boundaries, and downstream detector counts may include vehicles that were not part of the initial inflow. These ambiguities make it harder for the model to separate multiple uncertainty sources when training data are limited, reducing transfer performance.

\begin{figure}[!t]
    \centering
    \begin{subfigure}[b]{\textwidth}
        \centering
        \includegraphics[width=\textwidth]{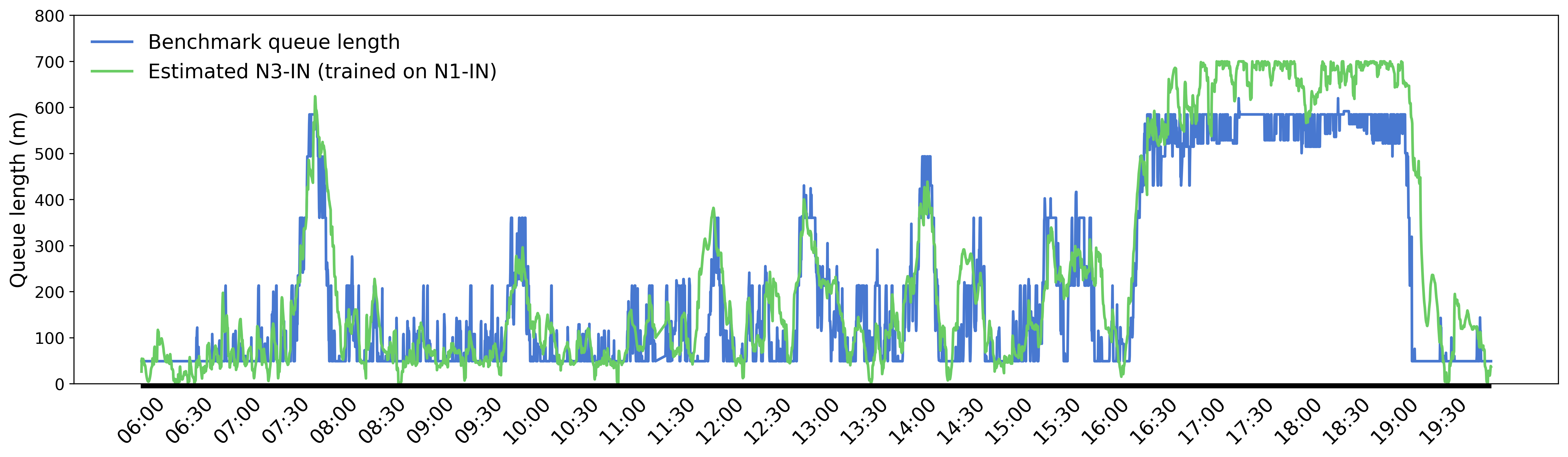}
        \caption{Model trained on \texttt{N1-IN} tested on \texttt{N3-IN} on 2023-11-14.}
    \end{subfigure}
    \begin{subfigure}[b]{\textwidth}
        \centering
        \includegraphics[width=\textwidth]{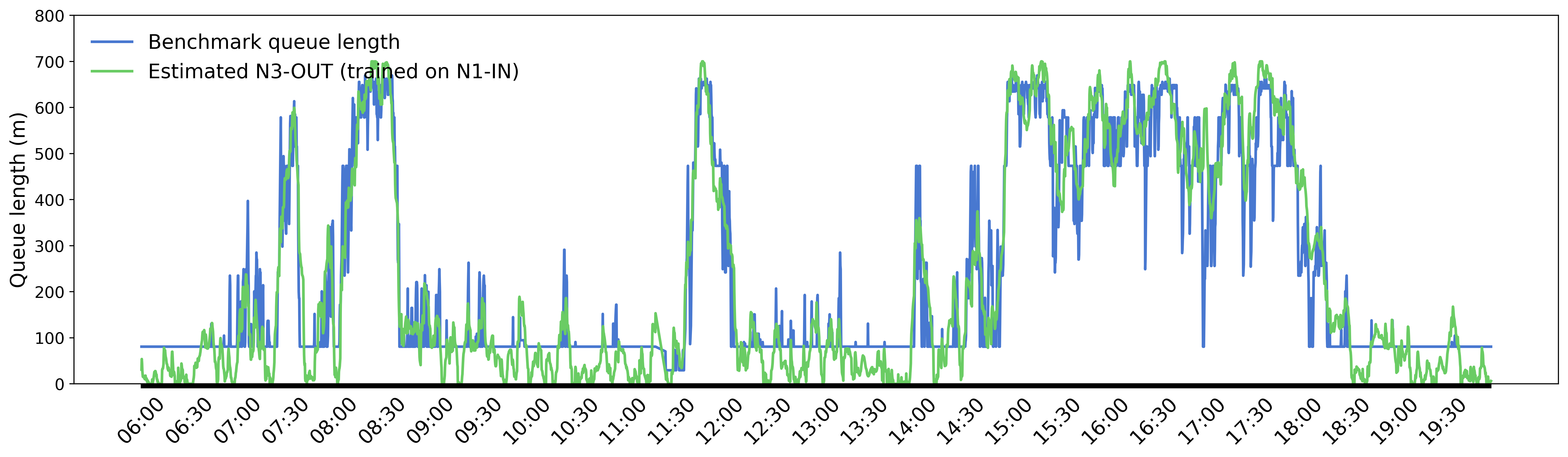}
        \caption{Q-Net trained on \texttt{N1-IN} tested on \texttt{N3-OUT} on 2023-11-14.}
    \end{subfigure}
    \caption{Queue length estimation for section \texttt{N3-IN} and \texttt{N3-OUT} using a model trained on \texttt{N1-IN} without fine-tuning (random seed 42).}
    \label{fig:visual_transferability}
\end{figure}

Figure~\ref{fig:visual_transferability} visualizes transfer results on \texttt{N3-IN} and \texttt{N3-OUT} for 2023-11-14 using Q-Net trained on \texttt{N1-IN}. The model captures multiple queue peaks and follows the benchmark during most queue formation and dissipation phases, demonstrating transferability to unseen sections and time periods. Two observations are worth noting. First, the benchmark queue never drops to 0 m, suggesting possible radar malfunction because continuous queues during early morning hours are unrealistic. Second, during stable afternoon queues, Q-Net sometimes overestimates the benchmark. From a traffic management perspective, this conservative bias is acceptable once the minimum queue needed to trigger control action is reached.

\subsection{Online estimation for real-time traffic control}
\label{sec:real_time}
The default Q-Net implementation estimates a fixed unobserved traffic flow rate using full-day cumulative net flow data (Subsection~\ref{sec:control_input}) and backward-aligns each aFCD timestamp to the preceding aggregation interval. This ex-post formulation is suitable for performance evaluation but not for causal real-time operation. To assess online applicability and isolate the effect of aFCD temporal delay, we implement two causal variants. The semi-online version estimates the unobserved traffic flow rate using only cumulative flow data available up to the current time, while retaining backward-aligned aFCD timestamps. The fully online version estimates the unobserved traffic flow rate in real time and uses each aFCD observation only after it becomes available. All other hyperparameters remain unchanged.

\begin{figure}[!t]
    \centering
    \includegraphics[width=\linewidth]{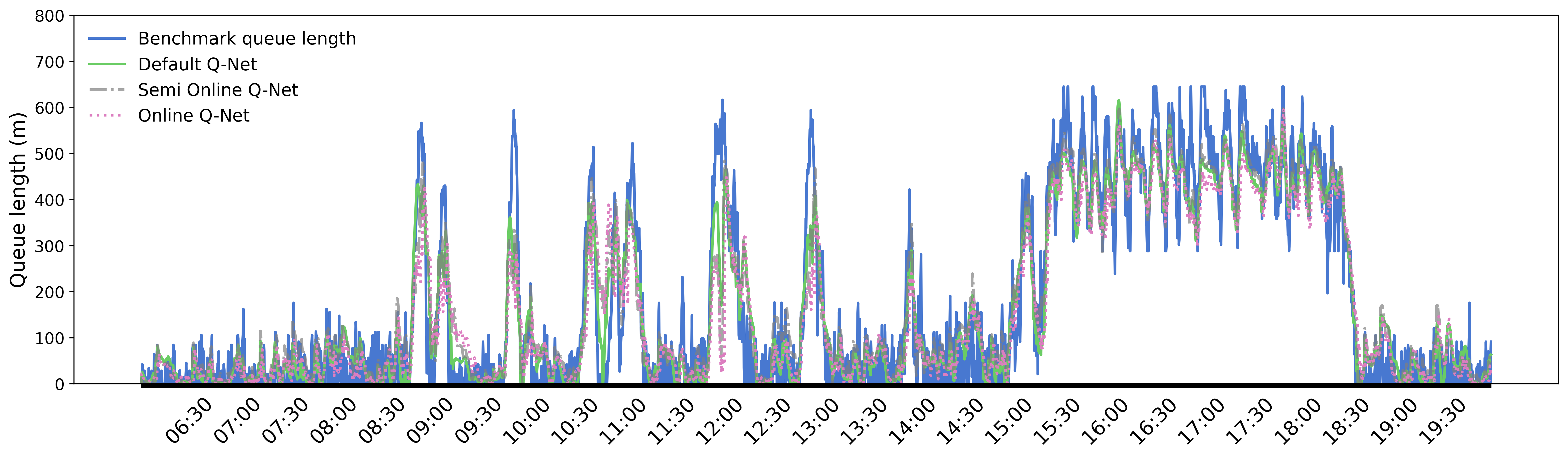}
    \caption{Comparison of queue length estimation for section \texttt{N1-IN} on 2023-11-22 (Wednesday) using default and online Q-Net. The default version estimates a fixed rate from full-day data via linear regression, while the online version updates the rate using data available up to the current time.}
    \label{fig:visual_realtime}
\end{figure}

Table~\ref{tab:real_time} and Figure~\ref{fig:visual_realtime} compare the default, semi-online, and fully online versions. The semi-online version performs similarly to the default Q-Net in terms of average metrics, indicating that causal estimation of the unobserved flow rate has limited impact. The fully online version shows lower all-day accuracy mainly because delayed aFCD availability amplifies the response delay during rapid queue transitions. However, during the afternoon peak, its performance remains close to the default Q-Net. This indicates that the impact of aFCD time lag is limited in periods where the queue state changes more gradually, whereas it becomes more pronounced during abrupt transitions. During rapid queue transitions, the default Q-Net reacts most promptly, the semi-online version shows a slightly longer delay, and the fully online version exhibits the largest delay. For example, during the queue formation and dissipation period around 08:30-09:30 on 2023-11-22, both online variants lag behind the default estimate but still follow the main trend. Overall, the online variants capture the dominant queue dynamics, suggesting that Q-Net remains promising for real-time monitoring and control applications where causal implementation is required.

\begin{table}[!t]
    \centering
    \caption{Performance comparison of Q-Net and its online version when trained and tested on \texttt{N1-IN}.}
    \begin{tabular}{c|ccc|ccc|ccc}
        \hline
        &\multicolumn{3}{c|}{\textbf{All day}}&\multicolumn{3}{c|}{\textbf{Morning peak}}&\multicolumn{3}{c}{\textbf{Afternoon peak}}\\
        &\textbf{RMSE} & \textbf{MAE} & \textbf{MAPE} &\textbf{RMSE} & \textbf{MAE} & \textbf{MAPE} &\textbf{RMSE} & \textbf{MAE} & \textbf{MAPE} \\\hline
\textbf{Default Q-Net}&86.77&58.14&56.22&78.62&47.62&74.64&88.36&69.25&17.41\\
\textbf{Semi-online Q-Net}&85.33&56.74&56.34&76.24&46.93&74.88&81.07&62.6&15.56\\
\textbf{Online Q-Net}&100.58&67.92&70.28&89.73&56.56&92.27&89.08&69.89&18.19
\\\hline
    \end{tabular}
    \label{tab:real_time}
\end{table}

\subsection{Ablation study}
\label{sec:abaltion_study}
 
\begin{figure}[!t]
    \centering
    \includegraphics[width=\linewidth]{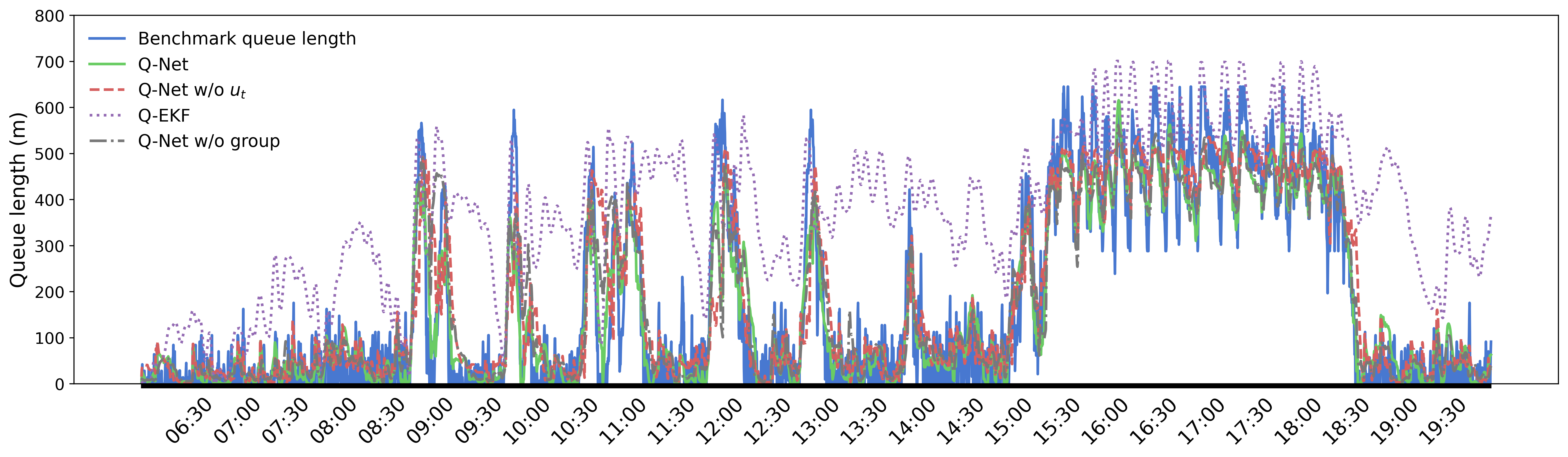}
    \caption{Queue length estimation results for different model variants on 2023-11-22 for section \texttt{N1-IN}.}
    \label{fig:ablation}
\end{figure}

As Q-Net's main components are the state-space formulation, learned Kalman gain, and grouped measurement strategy, we evaluate three ablation variants. The first, Q-Net w/o $u_t$, removes the control input from Equation~\eqref{eq:evolution}, yielding $x_t=x_{t-1}+w_t$. Since $u_t$ is derived from cumulative vehicle counts, this variant relies on aFCD only and tests the contribution of loop-detector-driven queue dynamics. The second variant, Q-EKF, replaces KalmanNet with a standard Extended Kalman Filter (EKF), isolating the benefit of learning the Kalman gain under nonlinear and time-varying uncertainty. The third variant, Q-Net w/o grouping, removes the local measurement grouping strategy in Equation~\eqref{eq:group_KF}, resulting in a location-specific model without spatial transferability. Results are shown in Table~\ref{tab:ablation} and Figure~\ref{fig:ablation}.

\medskip\noindent\textbf{State-space formulation and control input ($u_t$).} 
Comparing Q-Net w/o $u_t$ with the aFCD-only baselines in Table~\ref{tab:N1-IN} shows the value of the state-space formulation: even without loop-detector-derived queue changes, Q-Net w/o $u_t$ substantially improves over direct speed-based methods. Comparing Q-Net w/o $u_t$ with the default Q-Net then isolates the role of $u_t$. Without $u_t$, the model relies only on aFCD and is more affected by temporal lag, especially during queue formation around 11:30-13:00 in Figure~\ref{fig:ablation}. By incorporating flow-count-based queue changes $u_t$, the default Q-Net tracks dynamic buildup and dissipation more promptly.

Q-Net w/o $u_t$ performs slightly better during the afternoon peak in Table~\ref{tab:ablation}, when queues are long and relatively stable. In this condition, the true queue change is often close to zero, so the assumption $u_t\approx 0$ becomes reasonable, while the default model introduces minor performance degradation through noisy cumulative-count-based $u_t$. This does not diminish the overall value of the control input, which is more critical for queue formation and dissipation.

\medskip\noindent\textbf{Data-driven Kalman filtering.} 
Q-EKF performs worst across all metrics and fails to track queue formation and dissipation accurately (Figure~\ref{fig:ablation}). This highlights the importance of learning the Kalman gain from data. A standard EKF relies on predefined noise covariances, which are insufficient for the nonlinear and time-varying uncertainties observed in urban queue estimation.

\begin{table}[!t]
    \centering
    \caption{Performance comparison of model variants in the ablation study on section \texttt{N1-IN}.}
    \resizebox{\textwidth}{!}{%
    \begin{tabular}{c|ccc|ccc|ccc}
        \hline
        &\multicolumn{3}{c|}{\textbf{All day}}&\multicolumn{3}{c|}{\textbf{Morning peak}}&\multicolumn{3}{c}{\textbf{Afternoon peak}}\\
        &\textbf{RMSE} & \textbf{MAE} & \textbf{MAPE} &\textbf{RMSE} & \textbf{MAE} & \textbf{MAPE} &\textbf{RMSE} & \textbf{MAE} & \textbf{MAPE} \\\hline
        \textbf{Default}&86.77&58.14&56.22&78.62&47.62&74.64&88.36&69.25&17.41\\
        
        \textbf{Q-Net w/o $u_t$}&90.2&59.75&56.81&83.31&49.24&73.24&87.54&68.67&17.27\\
        
        \textbf{Q-EKF}&217.58&175.78&259.58&213.45&165.46&353.47&135.55&111.59&34.39\\

        \textbf{Q-Net w/o group}&90.12&60.26&58.0&88.1&51.4&83.91&89.78&71.3&16.63\\\hline
    \end{tabular}
    }
    \label{tab:ablation}
\end{table}

\medskip\noindent\textbf{Grouped measurement strategy.} Q-Net w/o group performs slightly worse than the default Q-Net (Table~\ref{tab:ablation}), even though it is trained as a location-specific model. This suggests that the grouping strategy does not impose an accuracy penalty for enabling spatial transferability. Instead, neighboring aFCD segments provide useful local context for gain estimation. The grouped design therefore supports spatial transferability (Subsection \ref{sec:transfer}) without compromising single-location performance.

\section{Discussion}
\label{sec:discussion}
Q-Net addresses a practical queue estimation setting in which traffic flows are only partially observed and direct queue measurements are unavailable during deployment. It uses two widely available and privacy-preserving data sources: loop detector counts and aggregated floating car data (aFCD). While
loop detector placement may differ across networks, our method remains applicable by accounting for the travel time between vehicle detection and arrival at the stop line, though performance may vary depending on detector placement and traffic conditions.

\noindent\textbf{Architectural considerations.} Q-Net uses loop detector data to derive queue change as the control input. This component is obtained through affine transformation and Fourier-based band-pass filtering, as described in Subsection~\ref{sec:control_input}. More specialized neural architectures could be designed for estimating queue changes, and we experimented with LSTM and attention-based alternatives. However, under limited training data, these models struggled to learn queue-change dynamics and noise uncertainties at the same time, and the current formulation performed better. A separate framework focused only on queue-change prediction and subsequent queue reconstruction is possible but beyond the scope of this study.

Q-Net introduces two practical deviations from classical Kalman filtering. First, it applies a projection operation to keep queue estimates within physically feasible bounds. Although this sacrifices some classical Kalman optimality, such ad hoc constraints are widely used in engineering practice to enforce physical feasibility \citep{simon2010kalman, prakash2014recursive} and performed well in our experiments. Second, Q-Net learns the Kalman gain with an unconstrained neural network, so the KalmanNet implementation does not enforce positive semi-definiteness of learned covariance surrogates. In traditional recursive estimation, violating this property may destabilize the filter and lead to divergence \citep{simon2006optimal}. Recent methods enforce positive definiteness using Cholesky decomposition \citep{ko2024cholesky}, but they require empirically tuned activation functions and more complex loss functions. We use an unconstrained architecture to retain representational flexibility and did not observe divergence or numerical instability during our experiments. Establishing formal stability guarantees for neural gain-learning filters remains an important direction for future research.

\noindent\textbf{Evaluation limitations.} The empirical evaluation has two main limitations. First, although the dataset is real-world, it covers only 16 days within a single month. Such data limitations are common in queue estimation studies \citep{car-01, car-05, car-08}. Simulation tools such as SUMO \citep{sengupta2021tqam} or AIMSUN \citep{abewickrema2025ensemble} could provide larger datasets, but simulating unobserved traffic flows in a high-fidelity manner is itself a complex research problem and was not pursued here.

Second, the spatial transferability evaluation is limited to \texttt{N1-IN}, \texttt{N3-IN}, and \texttt{N3-OUT}, which are located within the same corridor in Rotterdam due to data availability constraints. These sections differ in infrastructure layout, traffic direction, and unobserved traffic-flow patterns, allowing us to evaluate local spatial transferability across connected but non-identical sections. The shared corridor context may still limit the generality of the findings. Broader validation across more independent test settings, such as different corridors, network types, and cities, is therefore needed to assess whether the observed transferability generalizes beyond the local corridor context.

In this transfer setting, the learned gain adaptation mechanism is reused, while the unobserved net flow rate is derived from local loop detector data rather than fixed global parameters. However, because Q-Net treats each section as an aggregated storage unit, highly complex lane configurations may increase uncertainty in the effective net inflow estimation.

Advanced spatiotemporal architectures such as graph neural networks and Transformers could also be considered for transfer learning. However, defining spatial dependencies for such baselines is not straightforward in the current setup, where only one labeled training section is available and the target sections have different layouts and aFCD segmentations. More generally, rigorous benchmarking of such models is constrained by the limited availability of ground-truth queue length data across multiple locations, particularly for constructing graph relationships across multiple sections and training data-intensive models. Benchmarking Q-Net against these architectures under larger-scale transfer settings is therefore left for future work.

\noindent\textbf{aFCD penetration and data requirements.} The estimated aFCD penetration rate in our study region is approximately 18.5\% \citep{eisinga2025network}. Q-Net is designed to reduce reliance on high probe coverage because loop detector inputs drive the queue evolution model, and aFCD periodically corrects prediction drift through macroscopic average speeds. Previous studies suggest that penetration rates as low as 2-3\% can capture stable macroscopic speed trends \citep{herrera2010evaluation}. This does not directly guarantee accurate queue estimation at such low penetration rates, but it suggests that useful macroscopic information may remain available. Quantifying Q-Net's performance degradation under reduced aFCD penetration remains an important future task.

\noindent\textbf{Future extensions and emerging data.}
Q-Net represents queue length as a section-level macroscopic state. Lane-level heterogeneity, movement-specific blocking, and lane-specific spillback are not explicitly modeled. This choice is consistent with the available data, which consist of aggregated detector counts, segment-level aFCD speeds, and section-level radar benchmark queues. Future work could extend the framework to lane-resolved state estimation or multi-regime traffic representations for complex oversaturated intersections.

Such extensions may be supported by emerging high-resolution data sources, including lane-specific trajectories from Connected and Autonomous Vehicles (CAVs) and localized queue detections from camera-based computer vision systems. These complementary, high-granularity streams would provide the microscopic detail necessary to move beyond current macroscopic approximations.

\section{Conclusion}
\label{sec:conclusion}
This paper addresses the challenge of estimating queue length in partially observed environments by proposing Q-Net, a Kalman-based framework that integrates model-driven and data-driven principles. Evaluated on multiple main road sections in Rotterdam, Q-Net achieves competitive or improved performance compared with baseline methods and effectively mitigates the lag typically observed in aFCD-based estimates. The model’s strong performance across temporal, spatial, and real-time online settings validates our hybrid approach. These results demonstrate Q-Net's practicality as a robust solution for real-world deployment, offering accurate queue estimation without the need to maintain or install costly radar infrastructure.

\section*{CRediT authorship contribution statement}
\textbf{Ting Gao}: Writing – review \& editing, Writing – original draft, Visualization, Validation, Software, Methodology, Formal analysis, Data curation, Conceptualization. 
\textbf{Elvin Isufi}: Writing – review \& editing, Supervision, Software, Methodology, Conceptualization. 
\textbf{Winnie Daamen}: Writing – review \& editing, Supervision, Software, Methodology, Funding acquisition, Conceptualization.
\textbf{Erik-Sander Smits}: Resources, Methodology, Data curation, Conceptualization.
\textbf{Serge Hoogendoorn}: Writing – review \& editing, Supervision, Methodology, Funding acquisition, Conceptualization.

\section*{Acknowledgments}
The authors gratefully acknowledge the support of the Rotterdam municipality for providing the data. This research has been produced by the “EMERALDS” project which has received funding under the Horizon Europe research and innovation programme (GA No. 101093051).

\section*{Declaration of generative AI use}
During the preparation of this work, the authors used ChatGPT (OpenAI) and Gemini (Google) to assist with writing revision and language polishing. After using this tool, the authors reviewed and edited the content as needed and take full responsibility for the content of the publication.

\newpage
\appendix
\section{KalmanNet algorithm}
\label{sec:appendix_Kalmannet}
\begin{figure}[h]
    \centering
    \includegraphics[width=\linewidth]{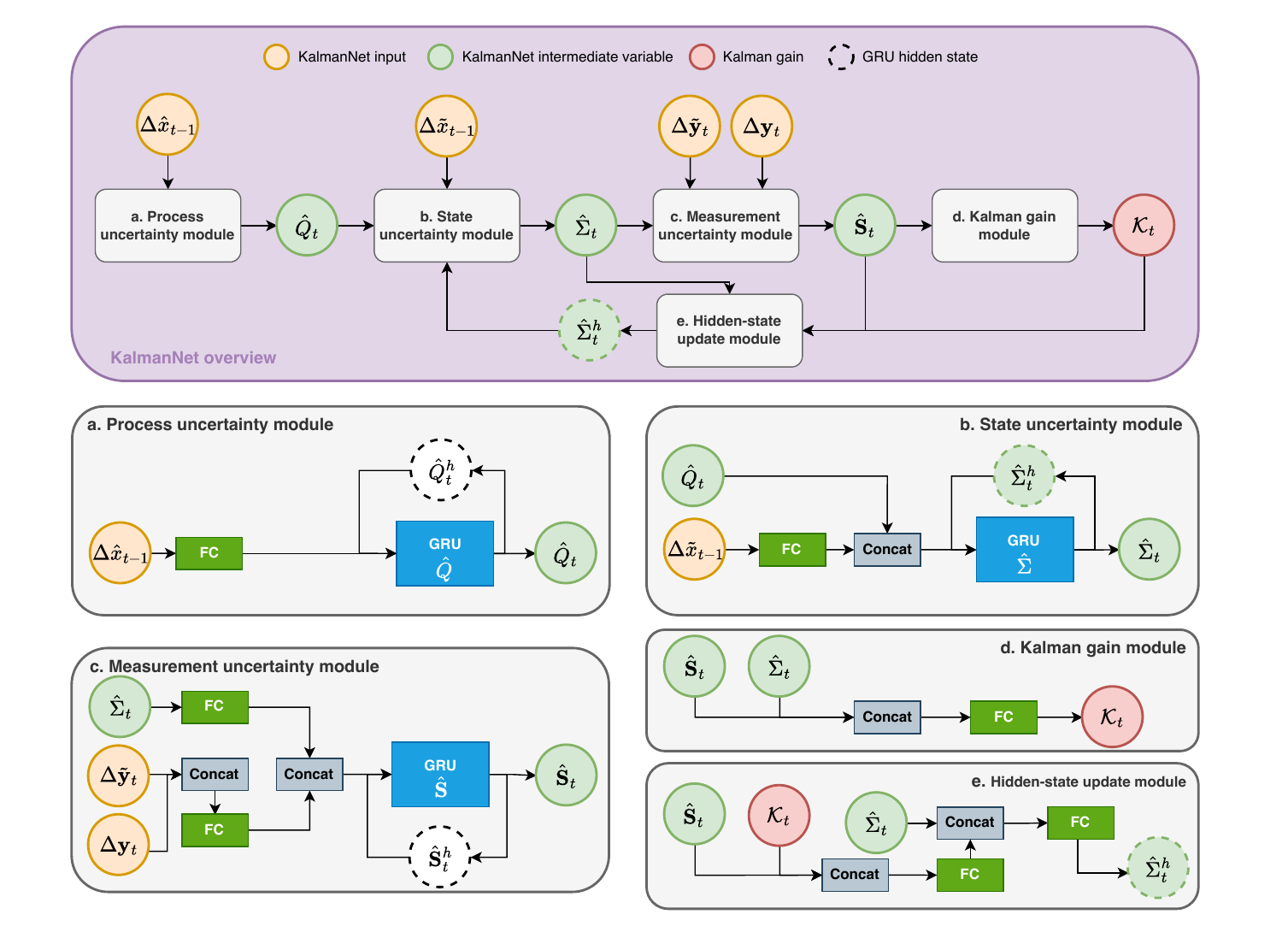}
    \caption{KalmanNet architecture for dynamic Kalman gain estimation. The top panel illustrates the overall KalmanNet pipeline, which takes in four temporal input features (\textit{forward evolution difference} $\Delta \tilde{x}_{t-1}$, \textit{forward update difference} $\Delta \hat{x}_{t-1}$, \textit{measurement difference} $\Delta \tilde{\mathbf{y}}_t$, \textit{innovation difference} $\Delta {\mathbf{y}}_t$)  to estimate the Kalman gain. Intermediate representations ($\hat{Q}_t, \hat{\Sigma}_t, \hat{\mathbf{S}}_t$ and $\hat{\Sigma}_t^h$) are computed through sequential submodules.
    The lower panels detail the submodules a, b, c, d, and e.
}
    \label{fig:KalmanNet_overview}
\end{figure}
Recall that $\Sigma_t$ and $\mathbf{S}_t$ denote the covariance matrices of the state $x_t$ and the measurements $\mathbf{y}_t$, while $Q_t$ and $\mathbf{R}_t$ represent the process and measurement noise covariances. In KalmanNet, we use the notation $\hat{(\cdot)}$ to denote learned, time-varying uncertainty representations inspired by these quantities, and ${(\cdot)}^h$ to denote internal latent states of recurrent modules. As shown in Figure~\ref{fig:KalmanNet_overview}, KalmanNet can be globally divided into five submodules: 

The \textbf{process uncertainty module} produces a time-varying representation $\hat{Q}_t$ based on the \textit{forward update difference} $\Delta \hat{x}_{t-1}$. Both $\hat{Q}_t$ and the \textit{forward evolution difference} $\Delta \tilde{x}_{t-1}$ are then fed into the \textbf{state uncertainty module} to produce the state-uncertainty embedding $\hat{\Sigma}_t$. Its associated latent stat $\hat{\Sigma}_t^h$ is later refined by the \textbf{hidden-state update module}. Next, $\hat{\Sigma}_t$, the \textit{innovation difference} $\Delta \mathbf{y}_t$, and the \textit{measurement difference} $\Delta \tilde{\mathbf{y}}_t$ are provided to the \textbf{measurement uncertainty module} to compute measurement-uncertainty embedding $\hat{\mathbf{S}}_t$. Finally, $\hat{\Sigma}_t$ and $\hat{\mathbf{S}}_t$ are passed to the \textbf{Kalman gain module}, which outputs the Kalman gain $\mathcal{K}_t$. All uncertainty-related matrices are represented in vectorized form throughout the network, which simplifies operations such as concatenation and allows the network to learn each entry without explicit matrix structure constraints.

\begin{enumerate}
    \renewcommand{\labelenumi}{\alph{enumi}.} 
    \item \textbf{Process uncertainty Module}: 
    The design is motivated by Equations~\eqref{eq:back_model_1} and~\eqref{eq:back_pred_1}, which suggest that evolution noise influences the predicted state estimate. The estimation follows:
    \begin{equation}
        \hat{Q}_t = \text{GRU}(\text{FC}(\Delta \hat{x}_{t-1})).
    \end{equation}
    The fully connected (FC) layers map the input into a higher-dimensional feature space. The gated recurrent unit (GRU) maintains a hidden state that is updated at each time step through its gating mechanism, allowing $\hat{Q}_t$ to reflect the complete history of \textit{forward update differences} rather than a static noise model.
    \item \textbf{State uncertainty Module}:
    This module estimates state-uncertainty embedding $\hat{\Sigma}_t$ using the learned process uncertainty representation $\hat{Q}_t$ (per Equation~\eqref{eq:back_pred_2}) and the \textit{forward evolution difference} $\Delta \tilde{x}_{t-1}$, which captures queue length evolution and system dynamics not reflected in process uncertainty alone:

    \begin{equation}
    \hat{\Sigma}_t, \hat{\Sigma}_t^h = \text{GRU}(\text{Concat}[\hat{Q}_t, \text{FC}(\Delta \tilde{x}_{t-1})]),
    \end{equation}
    where $\hat{\Sigma}_t^h$ denotes the GRU hidden state (later refined in module e) and Concat denotes concatenation. The GRU selectively retains relevant historical uncertainty patterns while discarding outdated information.
    
    \item \textbf{Measurement uncertainty module}: Following KalmanNet conventions, we denote the learned surrogate of the measurement covariance term by $\hat{\mathbf{S}}_t$. From Equation \eqref{eq:back_pred_4}, $\hat{\mathbf{S}}_t$ depends on the \textit{innovation difference} $\Delta \mathbf{y}_t$ (speed prediction errors) and $\hat{\Sigma}_t$ (queue length uncertainty). We additionally include the \textit{measurement difference} $\Delta \tilde{\mathbf{y}}_t$ to capture temporal variations in raw aFCD speed measurements, and the GRU gates control how past measurement patterns and current uncertainties influence the estimate:
    \begin{equation}
    \hat{\mathbf{S}}_t = \text{GRU}(\text{Concat}[\text{FC}(\hat{\Sigma}_t), \text{FC}(\text{Concat}[\Delta \tilde{\mathbf{y}}_t, \Delta \mathbf{y}_t])]).
    \end{equation}

    \item \textbf{Kalman gain module}: This module computes the Kalman gain $\mathcal{K}_t$ based on the queue length and speed measurement uncertainty embeddings, consistent with the inputs used by Kalman gain computation (state and measurement uncertainties):
    \begin{equation}
        \mathcal{K}_t = \text{FC}(\text{Concat}[\hat{\Sigma}_t, \hat{\mathbf{S}}_t]).
    \end{equation}
    The recursive nature of Kalman gain calculation (Equations \eqref{eq:back_pred_1}-\eqref{eq:back_update_2}), where each Kalman gain updates the state estimate and subsequently influences the next gain computation, is reflected through the GRU-based temporal dependencies in modules b and c.

    \item Hidden-state update: This module updates the internal hidden state $\hat{\Sigma}^h_t$ to retain temporal context across time steps, which is reflected by Equation \eqref{eq:back_update_2}:
    \begin{equation}
        \hat{\Sigma}^h_t = \text{FC}(\text{Concat}[\hat{\Sigma}_t, \text{FC}(\text{Concat}[\mathcal{K}_t, \hat{\mathbf{S}}_t])]).
    \end{equation}
\end{enumerate}
In summary, the Kalman gain is computed adaptively, leveraging both memory of past information and current predicted state and measurement data:
\begin{equation}
    \mathcal{K}_t =\text{KalmanNet}(\Delta \tilde{x}_t, \Delta \hat{x}_t, \Delta\tilde{\mathbf{y}}_t, \Delta {\mathbf{y}}_t).
    \label{eq:standard_KalmanNet}
\end{equation}
For additional implementation details, please refer to \citet{revach2022kalmannet}.

FC layers are typically followed by ReLU activations to produce non-negative feature representations. However, KalmanNet does not provide theoretical guarantees of positive semi-definiteness for the estimated covariance surrogates or strict non-negativity for scalar variance surrogates. In particular, outputs from the GRU layers are unconstrained, and the Kalman gain is computed via a two-layer network without explicit non-negativity enforcement on the final output. 

\section{Data split}
\label{appendix:data_split}
\begin{table}[H]
    \centering
    \caption{All data were collected in November 2023. The table presents the specific days assigned to the training, validation, and test sets in each split.}
    \begin{tabular}{cccc}
    \hline
    Split&Train days&Validation days&Test days\\
    \hline
    1 & 16, 17, 10, 27, 23, 13, 21, 15, 24, 25, 19 &20, 11 &14, 22, 18\\
    2 & 14, 22, 16, 17, 15, 24, 20, 21, 10, 18, 11 &23, 25 &27, 13, 19\\
    3 & 14, 24, 13, 23, 15, 17, 20, 10, 27, 19, 11&22, 18 & 21, 16, 25\\
    4 & 27, 20, 21, 23, 14, 17, 10, 22, 16, 19, 18 &13, 25 &15, 24, 11\\
    5 &20, 10, 22, 14, 16, 27, 15, 21, 13, 19, 11 &24, 18 &23, 17, 25\\
    6 &27, 22, 23, 24, 21, 15, 16, 17, 14, 11, 25&13, 19 &18, 10, 20 \\
    \hline
    \end{tabular}
\end{table}

\end{document}